%% file: eccv2022submissionCR.tex
\PassOptionsToPackage{table,xcdraw}{xcolor}
\documentclass[runningheads]{llncs}

\usepackage{graphicx}
\usepackage{tikz}
\usepackage{comment}
\usepackage{amsmath,amssymb,bm} 
\usepackage{color}
\usepackage{graphicx}
\usepackage{multirow}
\usepackage[ruled,vlined]{algorithm2e}
\usepackage{booktabs}

\definecolor{mypink3}{cmyk}{0, 0.7808, 0.4429, 0.1412/}
\definecolor{myblue3}{cmyk}{0, 0.1418, 0.9412, 0.3412/}

\usepackage[accsupp]{axessibility}  

\usepackage[pagebackref=true,breaklinks=true,letterpaper=true,colorlinks,bookmarks=false,citecolor=blue,linkcolor=blue,urlcolor=blue,]{hyperref}

\begin{document}
\pagestyle{headings}
\mainmatter
\def\ECCVSubNumber{4566}  
\def\ourAbbrname{$\mathbf{D^2ADA}$}
\def\ourFullname{\textbf{D}ynamic \textbf{D}ensity-aware \textbf{A}ctive \textbf{D}omain \textbf{A}daptation} 
\def\todo{\textcolor{red}{\textbf{[TODO]!}}}
\def\supp{supplementary material}
\def\ie{\textit{i.e.}}
\title{$\mathrm{D^2ADA}$: Dynamic Density-aware Active Domain Adaptation for Semantic Segmentation} 

\titlerunning{Dynamic Density-aware Active Domain Adaptation}
\authorrunning{T.-H. Wu et al.}
%
\author{Tsung-Han Wu\inst{1} \and Yi-Syuan Liou\inst{1} \and Shao-Ji Yuan\inst{1} \and Hsin-Ying Lee\inst{1} \\ Tung-I Chen\inst{1} \and Kuan-Chih Huang\inst{1} \and Winston H. Hsu\inst{1,2}}
\institute{$^1$National Taiwan University \qquad $^2$Mobile Drive Technology}
\maketitle

\begin{abstract}
    In the field of domain adaptation, a trade-off exists between the model performance and the number of target domain annotations. Active learning, maximizing model performance with few informative labeled data, comes in handy for such a scenario. In this work, we present \textbf{\ourAbbrname{}}, a general active domain adaptation framework for semantic segmentation. To adapt the model to the target domain with minimum queried labels, we propose acquiring labels of the samples with high probability density in the target domain yet with low probability density in the source domain, complementary to the existing source domain labeled data. To further facilitate labeling efficiency, we design a dynamic scheduling policy to adjust the labeling budgets between domain exploration and model uncertainty over time. Extensive experiments show that our method outperforms existing active learning and domain adaptation baselines on two benchmarks, GTA5 $\rightarrow$ Cityscapes and SYNTHIA $\rightarrow$ Cityscapes. With less than 5\% target domain annotations, our method reaches comparable results with that of full supervision. Our code is publicly available at \href{https://github.com/tsunghan-wu/D2ADA}{\color{blue}{https://github.com/tsunghan-wu/D2ADA}}.
\end{abstract}

\input{01_intro}

\input{02_relatedwork}

\input{03_method}

\input{04_experiment}

\input{05_conclusion}

\section*{Acknowledgement}
This work was supported in part by the Ministry of Science and Technology, Taiwan, under Grant MOST 110-2634-F-002-051, Mobile Drive Technology Co., Ltd (MobileDrive), and Industrial Technology Research Institute (ITRI). We are grateful to the National Center for High-performance Computing.

\clearpage
%
%
\bibliographystyle{splncs04}
\bibliography{egbib}

\clearpage

\input{supp}

\end{document}

%% file: 01_intro.tex
\section{Introduction}
\label{sec:intro}

Semantic segmentation is vital for many intelligent systems, such as self-driving cars and robotics. Over the past ten years, supervised deep learning methods \cite{chen2017deeplab,Chen2015SemanticIS,chen2017rethinking,chen2018encoder,lin2017refinenet,sun2019deep,wang2020deep,yuan2021segmentation,zhao2017pyramid} have achieved great success assisted by rich labeled datasets. However, obtaining large-scale datasets with manual pixel-by-pixel annotation is still costly in terms of time and effort. Thus, several prior works utilized \textit{domain adaptation} techniques to transfer the knowledge learned from the whole labeled source domain, such as simulated game environments, to real-world unlabeled target domain.

In the field of domain adaptation, a trade-off exists between model performance and the amount of target domain annotations. With sufficient target domain manual annotations, supervised learning method can reach high performance (e.g. $71.3$ mIoU on Cityscapes shown in Tab.~\ref{table:gta5_activ}). However, without any target labeling, the performance of unsupervised domain adaptation (UDA) methods \cite{chang2019all,cheng2021dual,du2019ssf,kim2020learning,liu2021bapa,luo2019CLAN,mei2020instance,shin2020two,tsai2018learning,vu2019advent,yang2020fda,zhang2021prototypical,zhou2021context,zou2018unsupervised,zou2019confidence} are still far below that of full supervision (e.g. $57.5$ mIoU on GTA \cite{richter2016GTA5}  $\rightarrow$ Cityscapes \cite{cordts2016cityscapes} reported by \cite{zhang2021prototypical}). Compared to the above two extreme cases, which are impractical in real-world applications, a more reasonable manner is to strike a balance between model performance and the cost of labeling efforts.

\begin{figure}[t]
    \centering
    \includegraphics[width=\linewidth]{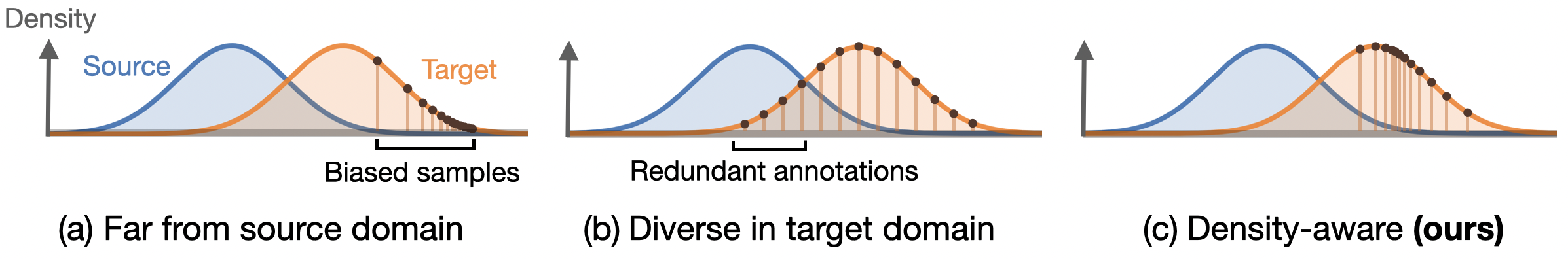}
    \caption{\textbf{Different Exploration Techniques in the ADA.} (a) \cite{su2020active,zhou2021discriminative} proposed acquiring labels of target samples that are far from the source domain by the trained domain discriminator. However, the selected biased samples are inconsistent with real target distribution. (b) \cite{fu2021transferable,prabhu2021active,singh2021improving} proposed selecting diverse samples in the target domain with clustering techniques for label acquisition. Nonetheless, redundant annotations exist on samples that are similar to the existing labeled source domain dataset. (c) We propose a \textbf{density-aware} ADA strategy that acquires labels for samples that are representative in the target domain yet scarce the source domain, which is better than only considering either the source domain (a) or the target domain (b).}
    \label{fig:fig1}
\end{figure}

Active learning, maximizing model performance with few informative labeled data, comes in handy for such a scenario. In the past few years, several works utilized model uncertainty \cite{gal2016dropout,roth2006margin,shin2021labor,siddiqui2020viewal,ijcnn,wang2016cost} or data diversity \cite{ash2020babdge,kirsch2019batchbald,ning2021multi,sener2018coreset,wu2021redal} as the indicator to select valuable samples for labeling. The core concept of \textit{uncertainty-based methods} is to acquire labels for data close to the model decision boundary, and the main idea of \textit{data diversity approaches} is to query labels for a batch of samples that are far away from each other in the feature space.




Recently, some studies leveraged uncertainty and diversity active learning methods for domain adaptation \cite{fu2021transferable,prabhu2021active,singh2021improving,su2020active,zhou2021discriminative}, named Active Domain Adaptation (ADA); however, these methods have two significant defects. First of all, existing techniques for exploring target domain distributions is not efficient. Fig.~\ref{fig:fig1} (a, b) shows two typical approaches. As observed in (a), selecting samples far from source domain distributions \cite{su2020active,zhou2021discriminative} might lead to outliers or biased data \cite{ovadia2019can}. As shown in (b), labeling diverse data in the target domain \cite{fu2021transferable,prabhu2021active,singh2021improving} might cause unnecessary annotations. In Fig.~\ref{fig:ALs}, we showed that the methods with 5\% annotations still performed worse than ours with 2.5x fewer labels.

\begin{figure}[t]
    \centering
    \includegraphics[width=\linewidth]{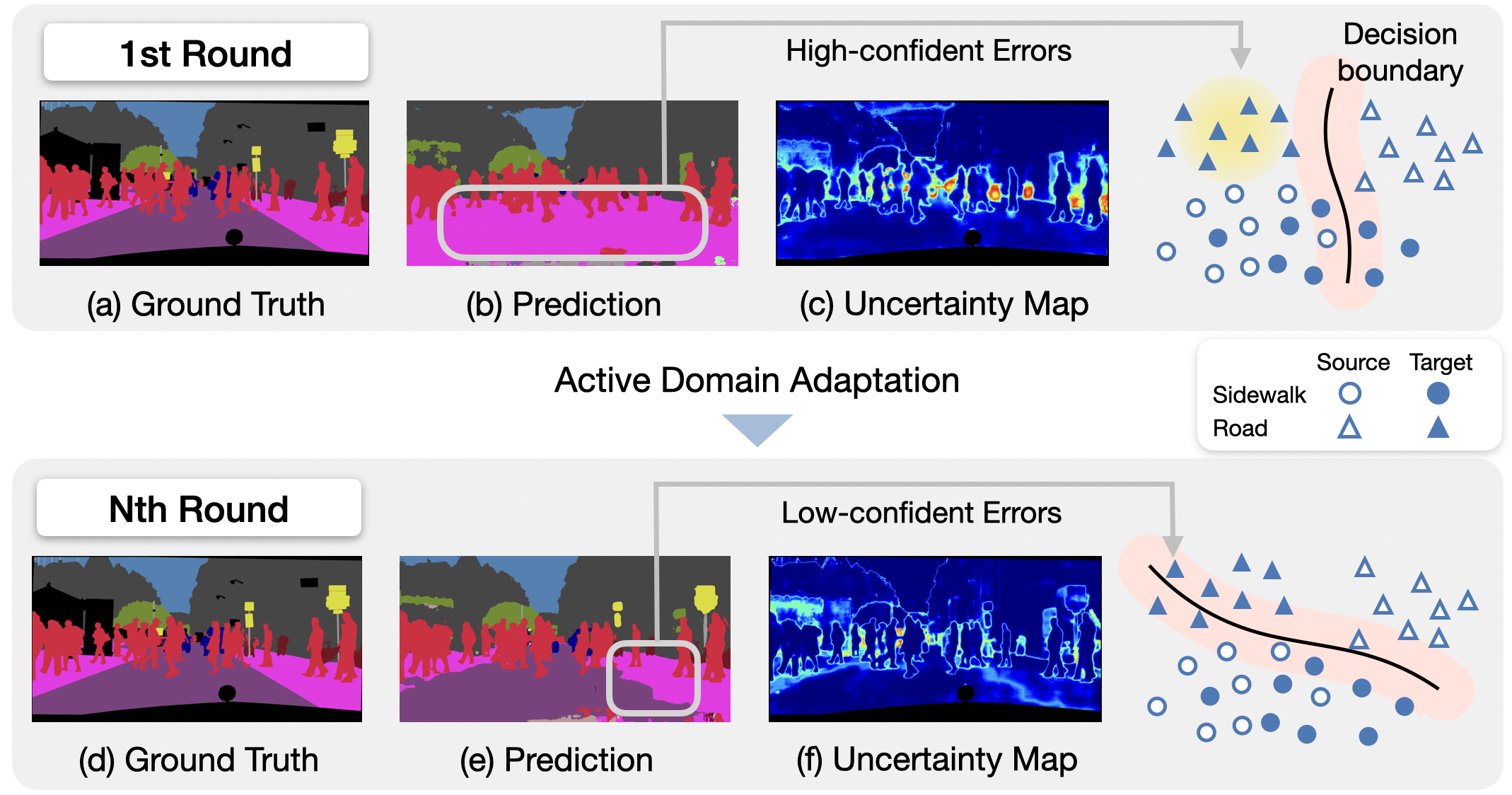}
    \caption{\textbf{Analysis of Uncertainty Criterion in the ADA.} Uncertainty-based active learning methods acquire labels of data close to the decision boundary (red background). Since its inability to detect high-confident error under severe domain shift, several representative samples in the target domain far from the decision boundary will not be selected (yellow background) in earlier rounds, resulting in low label efficiency under few labeling budgets as shown in (a, b, c). However, as the two domains gradually align by fine-tuning with acquired labels, the number of high-confident but erroneous regions are rapidly reduced. Meanwhile, uncertainty measurement is able to capture the low-confident error with an accurate model in later rounds, as shown in (d, e, f). Hence, we propose a dynamic scheduling policy to pay more attention on domain exploration in earlier stages and rely more on model uncertainty later (Sec. \ref{subsec:adaptive}).}
    \label{fig:motivation}
\end{figure}

Furthermore, in the field of active learning, label acquisition strategies will be executed for multiple rounds \cite{sener2018coreset,ijcnn}; yet, existing methods \cite{fu2021transferable,prabhu2021active,singh2021improving,su2020active,zhou2021discriminative}, combining uncertainty and diversity, did not notice the uncertainty measurement is less informative in the first few rounds in the ADA problem. As shown in Fig. \ref{fig:motivation} (a, b, c), model uncertainty might fail to detect high-confident but erroneous target domain regions (e.g. mispredict the pavement road as the sidewalk) under severe domain shift in earlier rounds. Thus, heavily relying on uncertainty cues led to poor results under low labeling budgets (see  Tab.~\ref{tab:Component Ablation} and Fig.~\ref{fig:ALs}).

To address the two problems, we present an \ourFullname{} (\textbf{\ourAbbrname{}}) framework for semantic segmentation. To select the most informative target domain data for labeling, we propose a novel \textit{density-aware selection method} to select data with the largest domain gaps. In this work, we use the term, \textbf{domain density}, to describe the prevalence of an observed sample in a specific domain. Then, we acquire the labels of regions with the largest density difference between the source and the target domain. As our intuition in Fig. \ref{fig:fig1} (c), since the model has already performed well on rich labeled source domain data, labeling few samples with high target domain density but low source domain density is sufficient for models to overcome the domain shift. Compared to prior works in Fig.~\ref{fig:fig1} (a, b), the superiority of our method is demonstrated in Fig.~\ref{fig:ALs}. Also, it is mathematically proved to reduce the generalization bound of domain adaptation.

To maximize the advantages of domain exploration and model uncertainty, we develop a \textit{dynamic scheduling policy} to adjust the labeling budgets between the two selection strategies over time. Since the key idea of the exploration technique is to reduce the domain gap, it greatly improves the performance in the first few rounds when the domain gap is large but is less effective later when the two domains are almost aligned as shown in Tab.~\ref{tab:Component Ablation}. In contrast, as discussed in Fig.~\ref{fig:motivation}, the strength of uncertainty is in later rounds (d, e, f), while its defect of ignoring high-confident but erroneous areas usually occurs in the earlier rounds (a, b, c). Based on the analysis, we designed a dynamic scheduling policy to allocate more labeling budgets for our density selection in earlier rounds and more for uncertainty selection later.

Experiments showed that our proposed \ourAbbrname{} surpasses existing active learning and domain adaptation baselines on two widely used benchmarks, GTA5 $\rightarrow$ Cityscapes and SYNTHIA $\rightarrow$ Cityscapes. Moreover, with less than 5\% target domain labels, our method reaches comparable results with full supervision.

To sum up, our contributions are highlighted as follows,
\begin{itemize}
    \itemsep=2pt
    \item We open up a new way for ADA that utilizes domain density as an active selection criterion.
    \item We design a dynamic scheduling policy to adjust the budgets between model uncertainty and domain exploration in the ADA problem.
    \item Extensive experiments demonstrate our method outperforms current state-of-the-art active learning and domain adaptation methods.
\end{itemize}



%% file: 02_relatedwork.tex
\section{Related Work}

We introduce domain adaptation methods for semantic segmentation with none or few target labels and briefly review the progress of deep active learning.
\subsection{Unsupervised Domain Adaptation for Semantic Segmentation}

Many researchers have delved into unsupervised domain adaptation (UDA) methods for semantic segmentation. One widely-used approach is to align the distribution of the source and target domain through adversarial training \cite{chang2019all,du2019ssf,luo2019CLAN,tsai2018learning,vu2019advent}. Another popular method is to apply self-training techniques  \cite{wang2021uncertainty,zhang2021prototypical,zheng2021rectifying,zhou2021context,zou2018unsupervised,zou2019confidence} by regularizing model uncertainty and assigning pseudo labels to reliable target domain data. 
However, the performance of distribution alignment is still not satisfying \cite{ning2021multi}, and self-training often suffers from label noise \cite{zhang2021prototypical}.
These defects make existing UDA methods hard to reach the performance of full supervision on labeled target domain data.

\subsection{Domain Adaptation for Semantic Segmentation with Few Target Labels}

To reduce the performance gap between UDA methods and supervision on the whole labeled target domain data, some recent works have 
explored the utilization of weak and few target domain annotations. 

\cite{paul2020domain} introduced a weakly-supervised domain adaptation (WDA) framework for semantic segmentation,
where image-level annotations are provided initially, and few manual or pseudo pixel-level labels are gradually added.

Some considered the semi-supervised domain adaptation (SSDA) scenario, where few pixel-level annotations are 
available during training.
\cite{wang2020alleviating} attempted to align the features of the source and target domain 
globally and semantically. \cite{chen2021semi} reduced the domain shift by mixing domains and knowledge distillation.


Unlike the above settings, where few labels were initially given, others used active learning strategies to acquire few target annotations from humans. For example, \cite{ning2021multi} chose diverse target domain images dissimilar to multiple source domain anchors for labeling. \cite{shin2021labor} utilized the inconsistency map generated by the disagreement of two classifiers to acquire few uncertain target pixel labels. 

Different from those prior works focusing on uncertain or diverse samples, our density-aware selection method acquires labels of few but vital samples that can bridge the gap between the source and target domains.

\subsection{Active Learning for Domain Adaptation} Active learning can reduce the annotation effort by gradually selecting the most valuable unlabeled data to be labeled. 

Conventional active learning strategies can be roughly divided into two categories: uncertainty-based methods and diversity-aware approaches. The main idea of uncertainty-based methods is to acquire labels of data close to the model decision boundary. For example, \cite{ijcnn,wang2016cost} utilized the softmax entropy, confidence, or margin of network outputs as the selection criterion. \cite{gal2016dropout,gal2017deep} proposed a better measurement of uncertainty with MC-dropout. On the other hand, the key concept of diversity approaches is querying annotations of a batch of samples far away from each other in the feature space. Prior works used clustering \cite{ash2020babdge} or greedy selection \cite{kirsch2019batchbald,sener2018coreset,wu2021redal} to achieve the diversity in a query batch.

In addition, several recent works have studied active domain adaptation scenarios that consider both uncertainty and diversity. \cite{su2020active,zhou2021discriminative} used the trained classifier and domain discriminator to select highly uncertain samples far from the source domain distribution. \cite{fu2021transferable,prabhu2021active,singh2021improving} utilized clustering to select highly uncertain samples that are diverse in target domain distribution. Very recently, \cite{xie2021active} proposed an energy-based approach and \cite{xie2022towards} presented the concept of regional impurity to address this problem.

To the best of our knowledge, we propose the first density-aware active domain adaptation strategy to acquire labels of representative samples with high density in the target domain but low density in the source domain. Also, we developed the first dynamic scheduling policy between domain exploration and model uncertainty for active domain adaptation.

%% file: 03_method.tex
\section{Method}
\label{sec:method}


\subsection{Overview}

In the Active Domain Adaptation (ADA) problem, we have a data pool of $C$-category semantic segmentation from two domains, comprising rich labeled data $D_S$ in the source domain and some data $D_T$ in the target domain. $D_T$ can be divided into $D_T^L$ and $D_T^U$, where $D_T^L$ contains the data in the target domain that has been labeled and $D_T^U$ stores the remaining unlabeled data.

Given the initial $D_T$ without any annotations, \ie, $D_T^L=\emptyset$, the ADA algorithm iteratively acquires annotations of few data in $D_T^U$ to maximize the performance improvement over the test data in the target domain. The label acquisition process will be performed several rounds and the labeling budgets in single round is a fixed number of pixels.


Our \ourFullname{} (\textbf{\ourAbbrname{}}) framework can be divided into 3 steps: (1) Train an initial model on $D_S \cup D_T$ with UDA methods as a warmup step. (2) Perform the density-aware selection to determine representative regions for each category according to the domain density and balance the budgets across classes. (Sec.~\ref{subsec:density}). (3) Dynamically determine the labeling budgets for domain exploration and model uncertainty in each round. Then, for the top-ranked regions in $D_T^U$, acquire ground truth labels and move them to $D_T^L$. Finally, fine-tune the model on $D_S \cup D_T^L$ in a supervised manner and go back to step (2) for a new round (Sec.~\ref{subsec:adaptive}). The complete and detailed algorithm can be referred in the \supp{}.

\subsection{Density-aware Selection}
\label{subsec:density}

To efficiently label the most informative target domain data, unlike prior domain exploration techniques causing biased samples (Fig.~\ref{fig:fig1}(a)) or redundant labeling (Fig.~\ref{fig:fig1}(b)), our density-aware approach (Fig.~\ref{fig:fig1}(c)) selects data representative in the target distribution but uncommon in the source distribution. To achieve this, we introduce the concept of \textbf{domain density} to calculate the prevalence of data in a single domain and estimate the categorical domain gaps between the two domains. By selecting data with the largest domain gaps, we can adapt the model to the target domain with the least possible amount of queried annotations.


In the following section, we first introduce the definition of domain density and provide ways to estimate it for each category. Then, we elaborate on our density difference metric and its theoretical foundation. Finally, we describe the motivation and operation of our class-balanced selection.


\smallskip\noindent{\bf Domain Density Estimation.}  Domain density is defined as a measurement of how common a region $R$ is predicted as a certain class in a particular domain.  Following prior works \cite{Casanova2020Reinforced,kasarla2019region,DBLP:conf/bmvc/MackowiakLGDLR18,siddiqui2020viewal}, we divide an image into multiple regions using the widely-used SLIC \cite{achanta2012slic} algorithm. To derive the domain density, we first calculate the feature $z$ of a region $R$ as the average feature over all the pixels within $R$. Similarly, a category $c \in [C] = \{1, 2, \cdots C\}$ is assigned to a region based on the predicted probability averaged across all pixels in the region. Given a domain $S$, a feature vector $z$, and a category $c$, we define the \textbf{source domain density} 

\begin{equation}
    d_S = p_S(z|c)
    \label{eq:density}
\end{equation}
as the probability that $z$ is classified as $c$ in $S$. To estimate $p_S(z|c)$, we construct the density estimators from all the observed $(z, c)$ pairs in the source domain dataset. The target domain follows the manner similarly.

In our implementation, a set of Gaussian Mixture Models (GMMs) are served as the density estimators. The construction of GMMs can be efficiently completed by offline and parallel execution, which takes about 0.01 seconds per region with an 8-core CPU personal computer. More discussions and details are left in the \supp{}. 

\smallskip\noindent{\bf Density Difference as Metric.} The key idea of our density-aware method is to select unlabeled target regions that contribute most to the domain gap so that we can adapt the model from the existing labeled source domain to the target domain efficiently. In practice, we use the difference between the target and the source domain density of a single region to evaluate the domain gap.

For the $i$-th unlabeled target region, we first obtain its source domain density $d_S^i$ and target domain density $d_T^i$ by feeding $(z_i , c_i)$ to the pre-constructed density estimators. Then, we introduce a metric to rank all regions in $D_T^U$:
\begin{equation}
    \pi_i =\log (\frac{d_{T}^i}{d_{S}^i}),
    \label{eq:diff}
\end{equation}
where $i = 1, 2, \cdots, |\text{\#regions}\in D_T^U|$. The ratio at the log scale is the log-likelihood difference of the two domain densities \cite{hastie2009elements}. According to Eq. \ref{eq:diff}, regions with larger $\pi$ are more prevalent in the target domain and less observable in the source domain. Therefore, we select regions with large $\pi$ values to replenish the insufficient knowledge of the source domain and to maximize the exploration of the target domain. For each category $c$, we sorted all unlabeled regions based on their importance scores. Then, a fixed number of top-ranked regions for each category will be selected for labeling until running out of annotation budgets.

\smallskip\noindent{\bf Theoretical Foundation.} To provide the theoretical foundations for our proposed density difference metric, we leverage the proposition presented in \cite{nguyen2022kl}
\begin{equation}
    \boldsymbol\ell_{test} \leq \boldsymbol\ell_{train} + \frac{M}{\sqrt{2}}\sqrt{D_{\text{KL}}(p_T(c, z)\ ||\ p_S(c, z))},
    \label{eq:theory}
\end{equation}
where $\boldsymbol\ell_{test}$ and $\boldsymbol\ell_{train}$ denote the testing and training loss respectively. $p_S$ and $p_T$ are the joint probability distributions of the category $c$ and the extracted feature $z$ in the source and the target domain distribution respectively. $M$ is a constant term and $D_{\text{KL}}(||)$ is the KL divergence of the two distributions.

As shown in Eq.~\ref{eq:theory}, to provide a tighter generalization bound for $\ell_{test}$, we aim to minimize the KL of the two joint probability distributions as follows:
\begin{multline}
  D_{\text{KL}}(p_T(c, z)\ ||\ p_S(c, z)) =D_{\text{KL}}(p_T(c)\ ||\ p_S(c))\\
  + {\mathbb{E}}_{p_T(c)} [D_{\text{KL}}(p_T(z|c)\ ||\ p_S(z|c))].
  \label{eq:proof}
\end{multline}
The proof of Eq. \ref{eq:proof} is provided in the \supp{}. \\

Since $p_T(c)$ and $p_S(c)$ are similar in most cases, we can obtain better generalization to the target domain by minimizing the expected value of the KL divergence from $p_S(z|c)$ to $p_T(z|c)$. Additionally, to deal with the label imbalance problem and to make the model perform well for each category for semantic segmentation, we opt to reduce the categorical KL term, \ie, $\ D_{\text{KL}}(p_T(z|c)\ ||\ p_S(z|c))$.

Although it is impossible to obtain the precise KL term by sampling all data points in the high dimensional continuous feature space, we could leverage the Monte Carlo simulation process to estimate the KL of the two distributions \cite{hershey2007approximating}. Given a sufficient number of $N$ observed region features $\{z_i\}_{i=1}^N$ predicted to a certain category $c$, the KL divergence of the two conditional probability distributions can be approximated as:

\begin{equation}
    D_{\text{KL}}(p_T(z|c)\ ||\ p_S(z|c))
  \approx \frac{1}{N} \sum_{i=1}^N \log( \frac{p_T(z_i|c)}{p_S(z_i|c)}) = \frac{1}{N} \sum_{i=1}^N \log( \frac{d_T^i}{d_S^i}) = \frac{1}{N} \sum_{i=1}^N \pi_i,
  \label{eq:KL}
\end{equation}
where $\pi$ is our designed metric to rank all regions in the same class (see Eq. \ref{eq:diff}).

Eq. \ref{eq:theory}, \ref{eq:proof}, \ref{eq:KL} bridge the theory and our proposed metric. By providing annotations for data that contribute the most to the KL divergence (large $\pi$ value) for each category, the domain shift between the source and target domain could be reduced, making the generalization bound of $\ell_{test}$ tighter.

\smallskip\noindent{\bf Class-balanced Selection.}
\label{subsubsec:category}
With our proposed density difference metric, the labeling budget in each round can be equally divided into $C$ categories. Nonetheless, we demonstrate that ADA can be achieved more efficiently.

We empirically observe the required budget for domain adaptation varies across classes, and thus it is inefficient to allocate the same budget to each class. For example, UDA methods can reach the performance close to fully supervised learning in some classes like ``vegetarian" and ``building". However, it is difficult for them to generalize to others like ``train" or ``sidewalk" (see Tab. \ref{table:gta5_activ}). Hence, we propose a class-balanced selection to fully utilize the labeling budget.

Intuitively, we would like to spend more budgets on hard categories, while fewer selection budgets for already well-aligned ones. According to Eq. \ref{eq:KL}, the average $\pi$ scores of a particular category can be regarded as an approximation of categorical KL divergence from the source domain to the target domain. Therefore, we use this indicator to balance the selection budget across classes. Suppose the total labeling budget of our density selection in each round is $B^d$ pixels, for a certain category $c$, the allocated budget $B^{d, c}$ is as follows:

\begin{equation}
    B^{d, c} = \frac{w_c}{\sum_{i \in [C]} w_i} \cdot B^d, \ \ w_c = \sigma(D_{KL}(p_T(z|c)\ ||\ p_S(z|c))).
    \label{eq:budget}
\end{equation}

In the Eq. \ref{eq:budget}, $\sigma$ is a normalization function to make the weighting term falls within a reasonable range. The equation ensures that more labels are assigned to classes with larger domain shift, while less are distributed to well-aligned ones.

\subsection{Dynamic Scheduling Policy}
\label{subsec:adaptive}

As described in Sec. \ref{sec:intro}, prior ADA practices acquired labels considering both uncertainty and domain exploration equally over time. However, these practices, not aware of the defects of both methods under different stages, caused poor labeling efficiency, especially under low labeling budgets in earlier rounds.

As shown in the comparison of Fig. \ref{fig:motivation} (a, b, c) and (d, e, f), uncertainty indicator is less effective under severe domain shift but provides informative cues when more target domain labeled data are acquired. Conversely, since the design concept of domain exploration is to conquer the domain shift, it is effective in the earlier rounds in ADA but the growth rapidly slows down in the later rounds as shown in Tab. \ref{tab:Component Ablation} (a, c). As a result, we develop a dynamic scheduling policy to benefit from both selection strategies and facilitate labeling efficiency.

Let $B$ be the constant labeling budgets  of each active selection round. In the $n$-th round, $B$ can be divided to $B^d_n$ for the density-aware approach and $B^u_n$ for an uncertainty-based method. Our budget allocation arrangements for the two approaches are as follows: 

\begin{equation}
\label{eq:lambda}
\begin{aligned}
& \quad \quad \quad \lambda = \alpha \cdot 2^{-\beta(n-1)}, \\
& B^d_n = \lambda B,\ B^u_n = (1-\lambda) B,
\end{aligned}
\end{equation}

where $\alpha$ is the balance coefficient between density-aware and the uncertainty-based method, and $\beta$ is the decay rate of the labeling budget of the density-aware approach. The reason for using half decay is to reflect the observation of the rapid domain shift reduction, which is discussed in Sec.~\ref{subsec:ablation}.
The above formula ensures that the density-aware method is heavily relied on when the domain gaps are large in the first few rounds, and the uncertainty-based method is mainly used when the gaps are diminished in the later rounds. The effectiveness of the designed policy is shown in Tab. \ref{tab:Component Ablation}. In our implementation, we use the conventional softmax entropy \cite{ijcnn} as the uncertainty measurement. 

After deciding the labeling budgets $B_n^d$ and $B_n^u$ for density-aware and uncertainty methods with our dynamic scheduling policy in each round, we select regions with the two selection strategies. Then, we acquire ground truth labels for these regions and move them from the unlabeled target set $D_T^U$ to the labeled target set $D_T^L$. Finally, we fine-tune the model with $D_S \cup D_T^L$ under full supervision and then execute the next active selection round.

%% file: 04_experiment.tex
\section{Experiments}

\subsection{Experimental Settings}

We elaborate the datasets, networks and active learning protocol in our experiments. The implementation details are reported in the \supp{}.

\smallskip\noindent{\bf Datasets.} We evaluated various active learning and domain adaptation methods on two widely-used domain adaptive semantic segmentation benchmarks: GTA5 $\rightarrow$ Cityscapes and SYNTHIA $\rightarrow$ Cityscapes.  Cityscapes \cite{cordts2016cityscapes} is a real-world self-driving dataset containing 2975 labeled training images, 500 labeled validation images and 1225 test images. GTA5 \cite{richter2016GTA5} is a synthetic dataset consisting of 24966 annotated images, which shares 19 semantic categories with Cityscapes. SYNTHIA dataset \cite{ros2016synthia} has 9400 synthetic annotated images, which share 16 semantic categories with Cityscapes. For a fair comparison, we evaluated all methods on the Cityscapes validation split.

\smallskip\noindent{\bf Models.} We used DeepLabV3+ \cite{chen2018encoder}, a semantic segmentation model, to evaluate all active learning strategies. To fairly compare our method with existing domain adaptation approaches, we report the results on both DeepLabV2 \cite{Chen2015SemanticIS} and DeepLabV3+ \cite{chen2018encoder}. The two models are based on the ResNet-101 \cite{he2016resnet} backbone.

\smallskip\noindent{\bf Active Learning Protocol.}  For all experiments, we first adopted \cite{tsai2018learning} to train the network with adversarial training as a warmup step. Then we performed $N$-round active selection comprising of the following steps: (1) Select a small number of regions with $B$ pixels from the unlabeled Cityscapes training split $D_T^U$ based on different active selection strategies. (2) Acquire the labels of these selected data and add them to $D^L_T$. (3) Fine-tune the model with $D_S \cup D_T^L$ in a supervised manner. In our setting, we choose $N=5$, $B=1\%$ of the total number of pixels in Cityscapes training split for both tasks.

\subsection{Comparison with Active Learning Baselines}
\label{subsec:ALresult}

We compared \ourAbbrname{} with 8 other active learning strategies, including random region selection (RAND), uncertainty-based methods (CONF \cite{ijcnn}, MAR \cite{ijcnn}, and ENT \cite{ijcnn}), hybrid methods (BADGE \cite{ash2020babdge} and ReDAL \cite{wu2021redal}), and existing Active Domain Adaptation baselines (AADA \cite{su2020active} and CLUE \cite{prabhu2021active}). The implementation details are described in the \supp{}.

\begin{figure*}[t]
    \centering
    \includegraphics[width=1.01\linewidth]{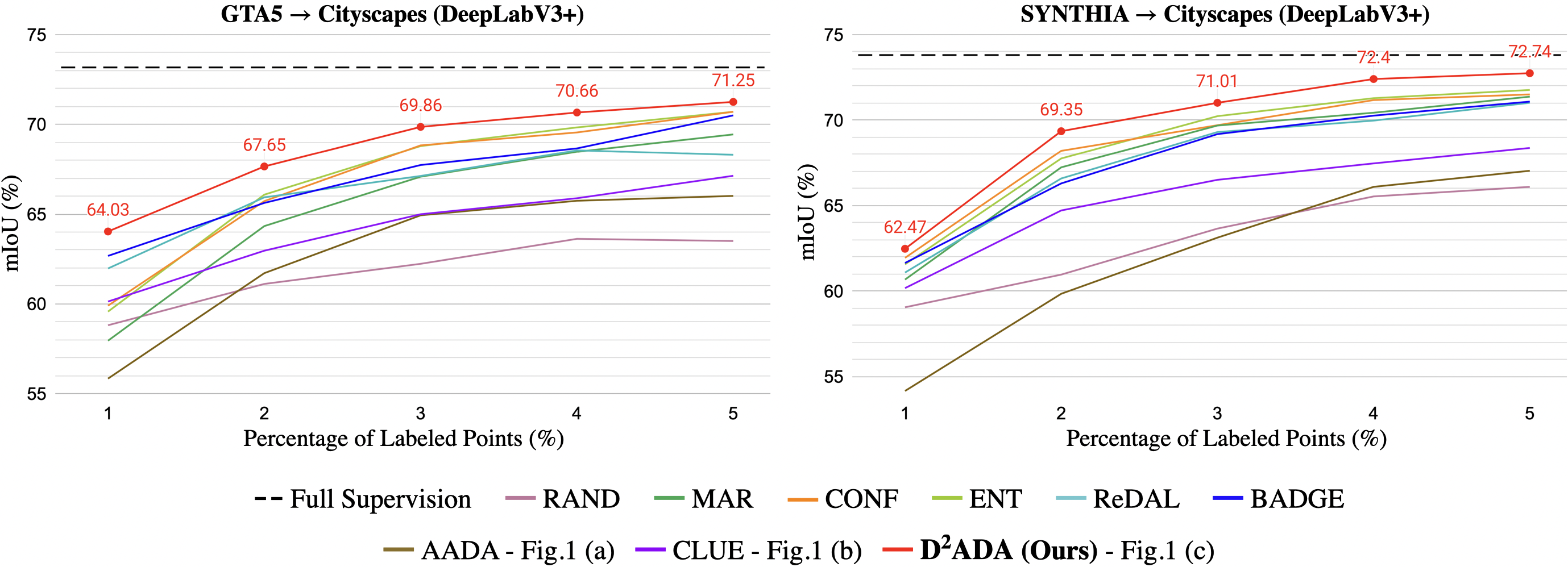}
    \caption{\textbf{Comparison with different active learning baselines.} On two widely-used benchmarks, our \ourAbbrname{} outperforms 8 existing active learning strategies, including uncertainty-based methods (MAR, CONF, ENT), hybrid approaches (ReDAL, BADGE), and ADA practices. The result suggested that compared to prior domain exploration methods shown in Fig. \ref{fig:fig1} (a, b), our density-aware method (c) achieves significant improvement. More explanations and observations are reported in Sec. \ref{subsec:ALresult}.}
    \label{fig:ALs}
\end{figure*}

The experimental results are shown in Fig.~\ref{fig:ALs}. In each subplot, the x-axis indicates the percentage of total target domain annotated points and the y-axis is the corresponding mIoU score achieved by the network.

\begin{table*}[t!]
\definecolor{LightCyan}{rgb}{0.88,1,1}
\renewcommand{\arraystretch}{1.2}
\begin{center}
\caption{\small \textbf{Comparison with different domain adaptation approaches on (a) GTA5 $\rightarrow$ Cityscapes and (b) SYNTHIA $\rightarrow$ Cityscapes.} Our proposed \ourAbbrname{} outperforms any existing methods on the overall mIoU and most per-class IoU with only 5\% target annotations with different network backbones. mIoU* in (b) denotes the averaged scores across 13 categories used in \cite{tsai2018learning}. To fairly compare all methods under the same number of annotations, we draw a chart in Fig.~\ref{fig:DAs}.}
\label{table:gta5_activ}
\centering
\resizebox{\textwidth}{!}{
\begin{tabular}{c l c c c c c c c c c c c c c c c c c c c|c}
\toprule
\multicolumn{21}{c}{(a) GTA5 $\to$ Cityscapes}\\
\midrule
\multirow{1}*{} & Method  & Road & SW & Build & Wall & Fence & Pole & TL & TS & Veg. & Terrain & Sky & PR & Rider & Car & Truck & Bus & Train & Motor & Bike & mIoU  \\
\midrule
\multirow{13}*{UDA} & 
AdaptSeg~\cite{tsai2018learning} & 86.5 & 36.0 & 79.9 & 23.4 & 23.3 & 35.2 & 14.8 & 14.8 & 83.4 & 33.3 & 75.6 & 58.5 & 27.6 & 73.7 & 32.5 & 35.4 & 3.9 & 30.1 & 28.1 & 42.4 \\
& ADVENT~\cite{vu2019advent} & 89.9 & 36.5 & 81.2 & 29.2 & 25.2 & 28.5 & 32.3 & 22.4 & 83.9 & 34.0 & 77.1 & 57.4 & 27.9 & 83.7 & 29.4 & 39.1 & 1.5 & 28.4 & 23.3 & 43.8\\
& SIMDA~\cite{wang2020differential} & 90.6 & 44.7 & 84.8 & 34.3 & 28.7 & 31.6 & 35.0 & 37.6 & 84.7 & 43.3 & 85.3 & 57.0 & 31.5 & 83.8 & 42.6 & 48.5 & 1.9 & 30.4 & 39.0 & 49.2\\


 & CBST~\cite{zou2018unsupervised} & 91.8 & 53.5 & 80.5 & 32.7 & 21.0 & 34.0 & 28.9 & 20.4 & 83.9 & 34.2 & 80.9 & 53.1 & 24.0 & 82.7 & 30.3 & 35.9 & 16.0 & 25.9 & 42.8 & 45.9 \\
& CRST~\cite{zou2019confidence} & 91.0 & 55.4 & 80.0 & 33.7 & 21.4 & 37.3 & 32.9 & 24.5 & 85.0 & 34.1 & 80.8 & 57.7 & 24.6 & 84.1 & 27.8 & 30.1 & 26.9 & 26.0 & 42.3 & 47.1 \\
& LTIR~\cite{kim2020learning} & 92.9 & 55.0 & 85.3 & 34.2 & 31.1 & 34.9 & 40.7 & 34.0 & 85.2 & 40.1 & 87.1 & 61.0 & 31.1 & 82.5 & 32.3 & 42.9 & 0.3 & 36.4 & 46.1 & 50.2\\
& FDA~\cite{yang2020fda} & 92.5 & 53.3 & 82.4 & 26.5 & 27.6 & 36.4 & 40.6 & 38.9 & 82.3 & 39.8 & 78.0 & 62.6 & 34.4 & 84.9 & 34.1 & 53.1 & 16.9 & 27.7 & 46.4 & 50.5 \\
& TPLD~\cite{shin2020two} & 94.2 & 60.5 & 82.8 & 36.6 & 16.6 & 39.3 & 29.0 & 25.5 & 85.6 & 44.9 & 84.4 & 60.6 & 27.4 & 84.1 & 37.0 & 47.0 & 31.2 & 36.1 & 50.3 & 51.2   \\
& IAST~\cite{mei2020instance}  & 93.8 & 57.8 & 85.1 & 39.5 & 26.7 & 26.2 & 43.1 & 34.7 & 84.9 & 32.9 & 88.0 & 62.6 & 29.0 & 87.3 & 39.2 & 49.6 & 23.2 & 34.7 & 39.6 & 51.5   \\
& ProDA~\cite{zhang2021prototypical}  & 87.8 & 56.0 & 79.7 & 46.3 & 44.8 & 45.6 & 53.5 & 53.5 & 88.6 & 45.2 & 82.1 & 70.7 & 39.2 & 88.8 & 45.5 & 59.4 & 1.0 & 48.9 & 56.4 & 57.5   \\
& CAMDA~\cite{zhou2021context} & 91.3 & 46.0 & 84.5 & 34.4 & 29.7 & 32.6 & 35.8 & 36.4 & 84.5 & 43.2 & 83.0 & 60.0 & 32.2 & 83.2 & 35.0 & 46.7 & 0.0 & 33.7 & 42.2 & 49.2 \\
& DPL-Dual~\cite{cheng2021dual} & 92.8 & 54.4 & 86.2 & 41.6 & 32.7 & 36.4 & 49.0 & 34.0 & 85.8 & 41.3 & 86.0 & 63.2 & 34.2 & 87.2 & 39.3 & 44.5 & 18.7 & 42.6 & 43.1 & 53.3 \\
& BAPA-Net~\cite{liu2021bapa}  & 94.4 & 61.0 & 88.0 & 26.8 & 39.9 & 38.3 & 46.1 & 55.3 & 87.8 & 46.1 & 89.4 & 68.8 & 40.0 & 90.2 & 60.4 & 59.0 & 0.0 & 45.1 & 54.2 & 57.4   \\
\midrule
\multirow{1}*{WDA} & 
WDA~\cite{paul2020domain} (Point) & 94.0 & 62.7 & 86.3 & 36.5 & 32.8 & 38.4 & 44.9 & 51.0 & 86.1 & 43.4 & 87.7 & 66.4 & 36.5 & 87.9 & 44.1 & 58.8 & 23.2 & 35.6 & 55.9 & 56.4  \\
\midrule
\multirow{1}*{SSDA} & 
ASS~\cite{wang2020alleviating} (+50 city) & 94.3 & 63.0 & 84.5 & 26.8 & 28.0 & 38.4 & 35.5 & 48.7 & 87.1 & 39.2 & 88.8 & 62.2 & 16.3 & 87.6 & 23.2 & 39.2 & 7.2 & 24.4 & 58.1 & 50.1  \\
\midrule
\multirow{5}*{ADA} & 
LabOR \cite{shin2021labor} (V2, 2.2\%) & 96.6 & 77.0 & 89.6 & 47.8 & 50.7 & 48.0 & 56.6 & 63.5 & 89.5 & 57.8 & 91.6 & 72.0 & 47.3 & 91.7 & 62.1 & 61.9 & 48.9 & 47.9 & 65.3 & 66.6 \\

& MADA \cite{ning2021multi} (V3+, 5\%) & 95.1 & 69.8 & 88.5 & 43.3 & 48.7 & 45.7 & 53.3 & 59.2 & 89.1 & 46.7 & 91.5 & 73.9 & 50.1 & 91.2 & 60.6 & 56.9 & 48.4 & 51.6 & 68.7 & 64.9   \\
& RIPU \cite{xie2022towards} (V3+, 5\%) & \textbf{97.0} & 77.3 & \textbf{90.4} & \textbf{54.6} & 53.2 & 47.7 & 55.9 & 64.1 & 90.2 & \textbf{59.2} & \textbf{93.2} & 75.0 & \textbf{54.8} & \textbf{92.7} & \textbf{73.0} & \textbf{79.7} & \textbf{68.9} & 55.5 & 70.3 & 71.2  \\
& \cellcolor{LightCyan}\ourAbbrname{} (V2, 5\%) & \cellcolor{LightCyan}96.3 & \cellcolor{LightCyan}73.6 & \cellcolor{LightCyan}89.3 & \cellcolor{LightCyan}50.0 & \cellcolor{LightCyan}52.3 & \cellcolor{LightCyan}48.0 & \cellcolor{LightCyan}56.9 & \cellcolor{LightCyan}64.7 & \cellcolor{LightCyan}89.3 & \cellcolor{LightCyan}53.9 & \cellcolor{LightCyan}92.3 & \cellcolor{LightCyan}73.9 & \cellcolor{LightCyan}52.9 & \cellcolor{LightCyan}91.8 & \cellcolor{LightCyan}69.7 & \cellcolor{LightCyan}78.9 & \cellcolor{LightCyan}62.7 & \cellcolor{LightCyan}57.7 & \cellcolor{LightCyan}71.1 & \cellcolor{LightCyan}69.7 \\
& \cellcolor{LightCyan}\ourAbbrname{} (V3+, 5\%) & \cellcolor{LightCyan}\textbf{97.0} & \cellcolor{LightCyan}\textbf{77.8} & \cellcolor{LightCyan}90.0 & \cellcolor{LightCyan}46.0 & \cellcolor{LightCyan}\textbf{55.0} & \cellcolor{LightCyan}\textbf{52.7} & \cellcolor{LightCyan}\textbf{58.7} & \cellcolor{LightCyan}\textbf{65.8} & \cellcolor{LightCyan}\textbf{90.4} & \cellcolor{LightCyan}58.9 & \cellcolor{LightCyan}92.1 & \cellcolor{LightCyan}\textbf{75.7} & \cellcolor{LightCyan}54.4 & \cellcolor{LightCyan}92.3 & \cellcolor{LightCyan}69.0 & \cellcolor{LightCyan}78.0 & \cellcolor{LightCyan}68.5 & \cellcolor{LightCyan}\textbf{59.1} & \cellcolor{LightCyan}\textbf{72.3} & \cellcolor{LightCyan}\textbf{71.3} \\
\midrule
\multirow{2}*{Target Only} & 
DeepLabV2 \cite{Chen2015SemanticIS} & 97.4 & 79.5 & 90.3 & 51.1 & 52.4 & 49.0 & 57.5 & 68.0 & 90.5 & 58.1 & 93.1 & 75.1 & 53.9 & 92.7 & 72.0 & 80.2 & 65.0 & 58.1 & 71.1 & 71.3 \\
& DeepLabV3+ \cite{chen2018encoder} & 97.6 & 81.3 & 91.1 & 49.8 & 57.6 & 53.8 & 59.6 & 69.1 & 91.2 & 60.5 & 94.4 & 76.7 & 55.6 & 93.3 & 75.8 & 79.9 & 72.9 & 57.7 & 72.2 & 73.2 \\
\bottomrule
\end{tabular}
}
\end{center}
\renewcommand{\arraystretch}{1.2}
\begin{center}
\centering
\resizebox{\textwidth}{!}{
\begin{tabular}{c l c c c c c c c c c c c c c c c c|c c}
\toprule
\multicolumn{19}{c}{(b) SYNTHIA $\to$ Cityscapes}\\
\midrule
\multirow{1}*{} & Method  & Road & SW & Build & Wall* & Fence* & Pole* & TL & TS & Veg. & Sky & PR & Rider & Car & Bus & Motor & Bike & mIoU & mIoU*\\
\midrule
\multirow{13}*{UDA} &
AdaptSeg \cite{tsai2018learning} & 79.2 & 37.2& 78.8& -& -& -& 9.9& 10.5& 78.2& 80.5& 53.5& 19.6& 67.0& 29.5& 21.6& 31.3&-  &45.9 \\
& ADVENT~\cite{vu2019advent} & 85.6& 42.2& 79.7& 8.7& 0.4& 25.9& 5.4& 8.1& 80.4& 84.1& 57.9& 23.8& 73.3& 36.4& 14.2& 33.0& 41.2 & 48.0 \\ 
& SIMDA\cite{wang2020differential} & 83.0 & 44.0 & 80.3 & - & - & - & 17.1& 15.8& 80.5& 81.8& 59.9& 33.1& 70.2& 37.3& 28.5& 45.8& - & 52.1 \\
& CBST~\cite{zou2018unsupervised}  & 68.0& 29.9& 76.3& 10.8& 1.4& 33.9& 22.8& 29.5& 77.6& 78.3& 60.6& 28.3& 81.6& 23.5& 18.8& 39.8& 42.6 & 48.9 \\
& CRST~\cite{zou2019confidence}  &67.7& 32.2& 73.9& 10.7& 1.6& 37.4& 22.2& 31.2& 80.8& 80.5& 60.8& 29.1 & 82.8& 25.0& 19.4& 45.3& 43.8 & 50.1 \\
& LTIR \cite{kim2020learning} & 92.6 & 53.2 & 79.2 & - & - & - & 1.6 & 7.5 & 78.6 & 84.4 & 52.6 & 20.0 & 82.1 & 34.8 & 14.6 & 39.4 & - & 49.3 \\
& FDA~\cite{yang2020fda} & 79.3 & 35.0 & 73.2 & - & - & - & 19.9 & 24.0 & 61.7 & 82.6 & 61.4 & 31.1 & 83.9 & 40.8 & 38.4 & 51.1 & - & 52.5 \\
& TPLD~\cite{shin2020two} & 80.9 & 44.3 & 82.2 & 19.9 & 0.3 & 40.6 & 20.5 & 30.1 & 77.2 & 80.9 & 60.6 & 25.5 & 84.8 & 41.1 & 24.7 & 43.7 & 47.3 & 53.5 \\
& IAST~\cite{mei2020instance} & 81.9 & 41.5 & 83.3 & 17.7 & 4.6 & 32.3 & 30.9 & 28.8 & 83.4 & 85.0 & 65.5 & 30.8 & 86.5 & 38.2 & 33.1 & 52.7 & 49.8 & 57.0 \\ 
& ProDA~\cite{zhang2021prototypical} & 87.8 & 45.7  & 84.6 & 37.1 & 0.6 & 44.0 & 54.6 & 37.0 & 88.1 & 84.4 & 74.2 & 24.3 & 88.2 & 51.1 & 40.5 & 45.6 & 55.5 & 62.0 \\
& CAMDA~\cite{zhou2021context} & 82.5 & 42.2 & 81.3 & - & - & - & 18.3 & 15.9 & 80.6 & 83.5 & 61.4 & 33.2 & 72.9 & 39.3 & 26.6 & 43.9 & - & 52.4 \\
& DPL-Dual~\cite{cheng2021dual} & 87.5 & 45.7 & 82.8 & 13.3 & 0.6 & 33.2 & 22.0 & 20.1 & 83.1 & 86.0 & 56.6 & 21.9 & 83.1 & 40.3 & 29.8 & 45.7 & 47.0 & 54.2 \\
& BAPA-Net~\cite{liu2021bapa} & 91.7 & 53.8 & 83.9 & 22.4 & 0.8 & 34.9 & 30.5 & 42.8 & 86.6 & 88.2 & 66.0 & 34.1 & 86.6 & 51.3 & 29.4 & 50.5 & 53.3 & 61.2 \\
\midrule
\multirow{1}*{WDA} &
WDA~\cite{paul2020domain} (Point) & 94.9 & 63.2 & 85.0 & 27.3 & 24.2 & 34.9 & 37.3 & 50.8 & 84.4 & 88.2 & 60.6 & 36.3 & 86.4 & 43.2 & 36.5 & 61.3 & 57.2 & 63.7 \\
\midrule
\multirow{1}*{SSDA} & 
ASS~\cite{wang2020alleviating} (+50 city) & 94.1 & 63.9 & 87.6 & - & - & - & 18.1 & 37.1 & 87.5 & 89.7 & 64.6 & 37.0 & 87.4 & 38.6 & 23.2 & 59.6 & - & 60.7  \\
\midrule
\multirow{4}*{ADA} &
MADA \cite{ning2021multi} (V3+, 5\%) & 96.5 & 74.6 & 88.8 & 45.9 & 43.8 & 46.7 & 52.4 & 60.5 & 89.7 & 92.2 & 74.1 & 51.2 & 90.9 & 60.3 & 52.4 & 69.4 & 68.1 & 73.3 \\
& RIPU\cite{xie2022towards} (V3+, 5\%) & \textbf{97.0} & \textbf{78.9} & 89.9 & 47.2 & 50.7 & 48.5 & 55.2 & 63.9 & \textbf{91.1} & 93.0 & 74.4 & 54.1 & \textbf{92.9} & \textbf{79.9} & 55.3 & 71.0 & 71.4 & 76.7 \\
& \cellcolor{LightCyan}\ourAbbrname{} (V2, 5\%) & \cellcolor{LightCyan}96.4 & \cellcolor{LightCyan}76.3 & \cellcolor{LightCyan}89.1 & \cellcolor{LightCyan}42.5 & \cellcolor{LightCyan}47.7 & \cellcolor{LightCyan}48.0 & \cellcolor{LightCyan}55.6 & \cellcolor{LightCyan}66.5 & \cellcolor{LightCyan}89.5 & \cellcolor{LightCyan}91.7 & \cellcolor{LightCyan}75.1 & \cellcolor{LightCyan}55.2 & \cellcolor{LightCyan}91.4 & \cellcolor{LightCyan}77.0 & \cellcolor{LightCyan}58.0 & \cellcolor{LightCyan}71.8 & \cellcolor{LightCyan}70.6 & \cellcolor{LightCyan}76.3 \\
& \cellcolor{LightCyan}\ourAbbrname{} (V3+, 5\%) & \cellcolor{LightCyan}96.7 & \cellcolor{LightCyan}76.8 & \cellcolor{LightCyan}\textbf{90.3} & \cellcolor{LightCyan}\textbf{48.7} & \cellcolor{LightCyan}\textbf{51.1} & \cellcolor{LightCyan}\textbf{54.2} & \cellcolor{LightCyan}\textbf{58.3} & \cellcolor{LightCyan}\textbf{68.0} & \cellcolor{LightCyan}90.4 & \cellcolor{LightCyan}\textbf{93.4} & \cellcolor{LightCyan}\textbf{77.4} & \cellcolor{LightCyan}\textbf{56.4} & \cellcolor{LightCyan}92.5 & \cellcolor{LightCyan}77.5 & \cellcolor{LightCyan}\textbf{58.9} & \cellcolor{LightCyan}\textbf{73.3} & \cellcolor{LightCyan}\textbf{72.7} & \cellcolor{LightCyan}\textbf{77.7} \\
\midrule
\multirow{2}*{Target Only} &
DeepLabV2 \cite{Chen2015SemanticIS} & 97.4 & 79.5 & 90.3 & 51.1 & 52.4 & 49.0 & 57.5 & 68.0 & 90.5 & 93.1 & 75.1 & 53.9 & 92.7 & 80.2 & 58.1 & 71.1 & 72.5 & 77.5 \\
& DeepLabV3+ \cite{chen2018encoder} & 97.6 & 81.3 & 91.1 & 49.8 & 57.6 & 53.8 & 59.6 & 69.1 & 91.2 & 94.4 & 76.7 & 55.6 & 93.3 & 79.9 & 57.7 & 72.2 & 73.8 & 78.4 \\
\bottomrule
\end{tabular}
}
\end{center}
\end{table*}
The improvement of our proposed \ourAbbrname{} is significant compared to prior active learning strategies. Under 5\% labels, \ourAbbrname{} surpassed the second-placed approach by 0.55 mIoU in GTA $\rightarrow$ Cityscapes, whereas the gap between the second and fourth place was only 0.2. Similarly, in SYNTHIA $\rightarrow$ Cityscapes, \ourAbbrname{} outperformed second place by 0.98 mIoU, while the difference between the second and fourth place was 0.39. Moreover, our method consistently prevailed over other methods by more than 0.5 mIoU under all cases. The results verify our method makes huge progress in active domain adaptation problems.

On the GTA5 $\rightarrow$ Cityscapes, our proposed \ourAbbrname{} surpasses uncertainty-based active learning methods  (ENT, CONF, and MAR) for over 3\% mIoU at the first round. This suggests that
density as a selection metric clearly benefits ADA over uncertainty against domain shift in the first few rounds.

Compared with existing ADA methods (AADA, CLUE) or hybrid active learning approaches (ReDAL, BADGE), our method also achieves better performance under any tasks and with any amount of budget. On the two tasks, AADA even performs worse than random selection in the early stage. We conjecture that the selected biased samples might be inconsistent with the real target distribution as shown in Fig.\ref{fig:fig1} (a). As for CLUE, the methods with 5\% annotations still performed worse than our method with merely 2.5x fewer labels (2\% annotations) on the two tasks. We infer the main reason might be that they only considered choosing diverse target domain data but did not avoid the selection that were likely to appear in the source distribution as observed in Fig.\ref{fig:fig1} (b).

\subsection{Comparison with Domain Adaptation Methods}

We compared our \ourAbbrname{} with various domain adaptation methods, including unsupervised domain adaptations (UDAs) \cite{cheng2021dual,kim2020learning,liu2021bapa,mei2020instance,shin2020two,tsai2018learning,vu2019advent,wang2020differential,yang2020fda,zhang2021prototypical,zhou2021context,zou2018unsupervised} weakly-supervised domain adaptation (WDA) \cite{paul2020domain}, semi-supervised domain adaptation (SSDA) \cite{chen2021semi,wang2020alleviating} and active domain adaptations (ADA) \cite{ning2021multi,shin2021labor,xie2022towards,xie2021active}\footnote{We report the official results of the UDA, WDA, SSDA and ADA methods and train the DeepLabV2 and DeepLabV3+ ourselves.}.

Tab. \ref{table:gta5_activ} (a, b) shows the result of GTA5 $\rightarrow$ Cityscapes and SYNTHIA $\rightarrow$ Cityscapes respectively. Compared with the current state-of-the-art UDA method \cite{zhang2021prototypical}, our method achieves an advantage of more than 10\% mIoU with only 5\% annotation effort. Furthermore, the performance of our method can achieve over 97\% of the result of full supervision on target labeled data on both DeepLabV2 and DeepLabV3+ models. On the two tasks, compared with other label-efficient approaches (WDA, SSDA, ADA), our method not only performs better on the overall mIoU score but also gains improvement on small objects, such as ``traffic sign (TS)" or ``bicycle", and difficult categories, like ``fence" and ``pole".



\begin{figure}[t]
    \centering
    \includegraphics[width=0.9\linewidth]{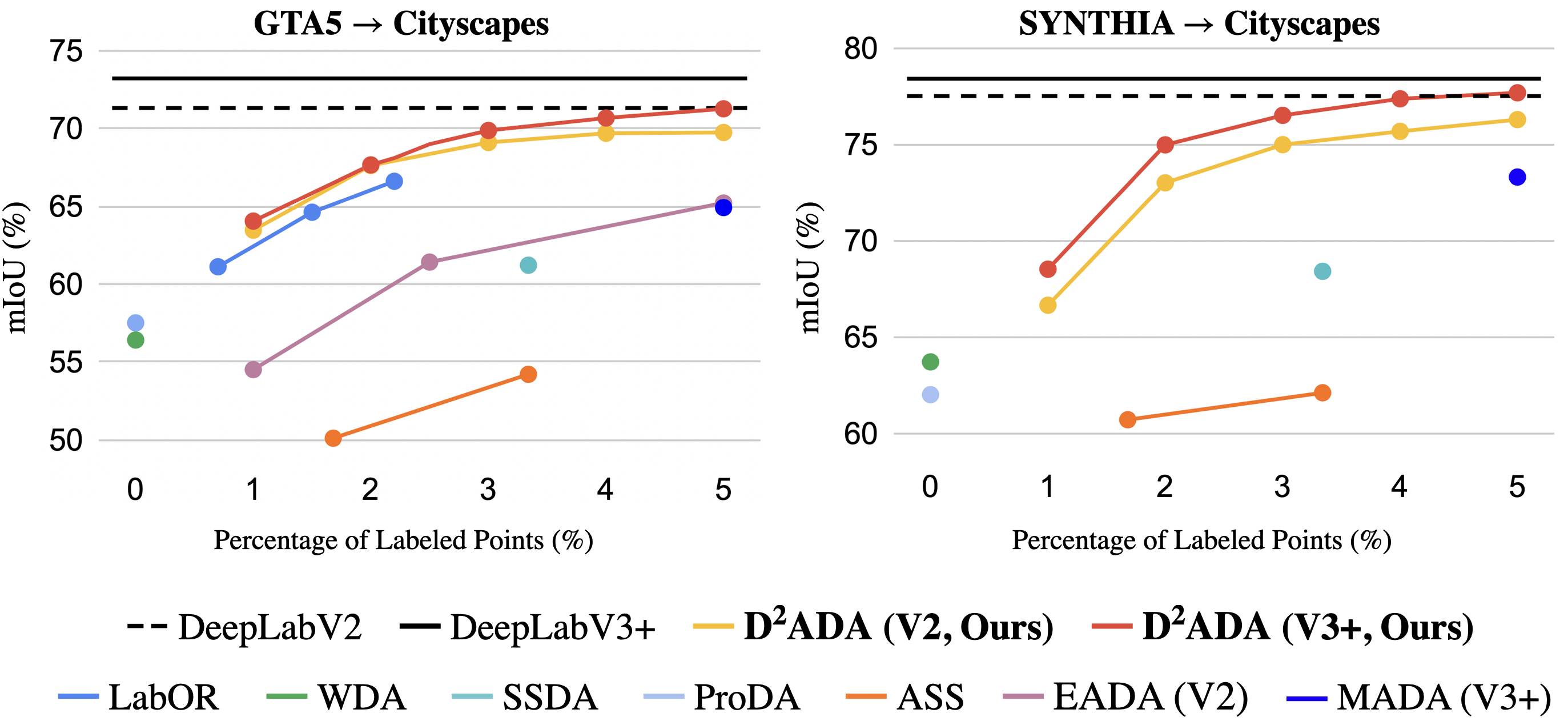}
    \caption{\textbf{Comparison with various label-efficient domain adaptation methods.} We compare our method with the state-of-the-art UDA method, ProDA \cite{zhang2021prototypical} as well as various label-efficient approaches, including WDA \cite{paul2020domain}, SSDA \cite{chen2021semi}, ASS \cite{wang2020alleviating}, MADA \cite{ning2021multi}, LabOR \cite{shin2021labor} and EADA \cite{xie2021active}. On two tasks and networks, our proposed \ourAbbrname{} achieves the best result under the same number of labels.}
    \label{fig:DAs}
\end{figure}

Since the labeling budgets used in prior label-efficient methods differ, Fig.~\ref{fig:DAs} makes a fair comparison of these methods. The x-axis indicates the percentage of total target domain annotated points and the y-axis is the corresponding mIoU score achieved by the network. Obviously, our approach achieves higher performance over existing methods with the same number of labeled pixels. 

Furthermore, according to statistics, our selected regions contain only 3.1 different categories on average, with 50\% regions less than two categories. Thus, compared with prior works selecting scattered pixels~\cite{shin2021labor} or small regions~\cite{xie2022towards}, our method, selecting superpixel-level regions, are more friendly for annotators to label (by drawing few polygons) and would require much less labeling effort.



\newcommand{\tabincell}[2]{\begin{tabular}{@{}#1@{}}#2\end{tabular}} 
\begin{table}[!t]
\caption {\textbf{Ablation studies.} The three columns on the right show the model performance when 1\%, 3\%, and 5\% target annotations are acquired by different active selection strategies. The results validate the effectiveness of our density-aware method, class-balanced selection, and dynamic scheduling policy as the discussion in Sec.~\ref{subsec:ablation}. It also demonstrates uncertainty perform worse with low labeling budgets (1\%, 3\%).}
\centering
\small
\setlength\tabcolsep{2pt}{
\begin{tabular}{c|cccc|ccc}
\toprule 
\multicolumn{1}{c|}{} & \multicolumn{4}{c|}{Components} & \multicolumn{3}{c}{mIoU wrt. labels} \\
\midrule
& \tabincell{c}{\footnotesize{uncertainty}} & \tabincell{c}{\footnotesize{density}\\\footnotesize{aware}} & \tabincell{c}{\footnotesize{class}\\\footnotesize{balance}} &  \tabincell{c}{\footnotesize{dynamic}\\\footnotesize{scheduling}} & \tabincell{c}{1\%} & \tabincell{c}{3\%} & \tabincell{c}{5\%}\\
\midrule
(a) & \checkmark & &  & & 59.97 & 68.79 & 70.70\\
(b) &  & \checkmark &  & &  62.75 & 68.81 & 69.82\\
(c) &  &\checkmark & \checkmark  & & \textbf{64.03} & 69.38 & 70.69\\
(d) & \checkmark&\checkmark &  \checkmark &  &63.30 & 69.66 & 71.15\\
(e) & \checkmark&\checkmark & \checkmark & \checkmark & \textbf{64.03} &  \textbf{69.86} & \textbf{71.25} \\

\bottomrule  
\end{tabular}%
}
\label{tab:Component Ablation}%

\end{table}%

\subsection{Ablation Studies}
\label{subsec:ablation}

We also conduct experiments to validate the effectiveness of all the proposed components with DeepLabV3+ backbone on GTA5 $\rightarrow$ Cityscapes. Extensive experiments about hyper-parameter settings and other in-depth analyses are left in the \supp{}.

Tab. \ref{tab:Component Ablation} demonstrates the effectiveness of our density-aware selection, class-balanced selection, and dynamic scheduling policy. First, from the comparison of (a, b), with low budgets (1\%, 3\%), our density-aware selection solves the defect of uncertainty metrics, echoing our motivations stated in Fig.~\ref{fig:motivation}.

\begin{figure*}[!h]
    \centering
    \includegraphics[width=0.85\linewidth]{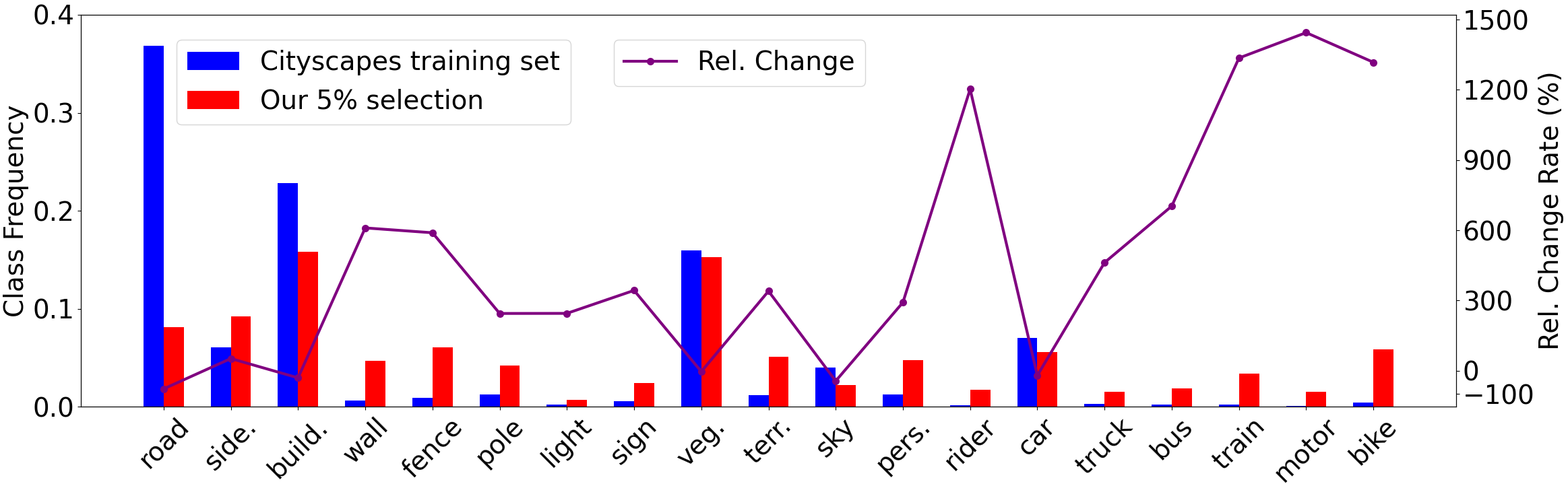}
    \caption{\textbf{The label distribution of our selected regions.} Following \cite{xie2022towards}, we draw histograms to compare the label distribution of the original target domain data (blue) and our 5\% selected regions (red). The purple line chart shows the relative changes from the target dataset to our selection. Evidently, our class-balanced selection acquires more labels on minor classes, like 12x more labels on ``train", ``motor", and ``bike".}
    \label{fig:vis}
\end{figure*}

Second, as shown in Fig.~\ref{fig:vis}, due to the class-balanced selection, our method queries more labels on minor classes but fewer labels on well-learned classes. Tab.~\ref{tab:Component Ablation} (b, c) also shows our class-balanced selection boosts the performance by over 0.5 mIoU in all cases. These results suggest that spending more labeling budgets on hard categories but few budgets on well-aligned ones is effective.

Third, as shown in Tab.~\ref{tab:Component Ablation} (a, e) and (c, e), given any budget, our dynamic scheduling policy surpasses the density or uncertainty methods alone. 
Also, the comparison of (d) and (e) validates our policy outperforms the ``equally distributed" budget allocation under all situations. The results indicate that our dynamic scheduling strategy can take advantage of the two methods, further improving the labeling efficiency.

%% file: 05_conclusion.tex
\section{Conclusion}

We propose \ourAbbrname{}, a novel active domain adaptation method for semantic segmentation. By labeling samples with high probability density in the target domain yet with low probability density in the source domain, models can conquer the domain shift with limited annotation effort. Furthermore, we design the first dynamic scheduling policy to balance labeling budgets between domain exploration and model uncertainty for the ADA problem.

%% file: supp.tex
\appendix
\def\etal{\textit{et al. }}
\def\ie{\textit{i.e.}}
\begin{center}
\LARGE{\textbf{Supplementary Material}}
\end{center}

\section{Proof}
\label{sec:proof}
We would like to prove Eq.~\ref{eq:proof} in Sec.~\ref{subsec:density} of the main paper.
\begin{equation*}
\small{
\begin{aligned}
& D_{\text{KL}}(p_T(c, z)\ ||\ p_S(c, z)) \\
& \quad = {\mathbb{E}}_{p_T(c, z)}[\log p_T(c, z) - \log p_S(c, z)]\\
& \quad = {\mathbb{E}}_{p_T(c, z)}[\log p_T(c) + \log p_T(z|c) - \log p_S(c) - \log p_S(z|c)] \\
& \quad = {\mathbb{E}}_{p_T(c, z)}[\log p_T(c) - \log p_S(c)] \\
& \quad \quad \quad \quad \quad + {\mathbb{E}}_{p_T(c, z)} [\log p_T(z|c) - \log p_S(z|c)]\\
& \quad = {\mathbb{E}}_{p_T(c)}[\log p_T(c) - \log p_S(c)]\\
& \quad \quad \quad \quad \quad + {\mathbb{E}}_{p_T(c)} [{\mathbb{E}}_{p_T(z|c)} [\log p_T(z|c) - \log p_S(z|c)]]\\
& \quad = D_{\text{KL}}(p_T(c)\ ||\ p_S(c)) + {\mathbb{E}}_{p_T(c)} [D_{\text{KL}}(p_T(z|c)\ ||\ p_S(z|c))]
\end{aligned}
}
\end{equation*}

\section{Implementation Details}
\label{sec:impl}

We conducted all experiments on an 8-core CPU personal computer with an NVIDIA RTX3090 GPU. The following section elaborates on the implementation details of UDA warm-up, density-aware selection, dynamic scheduling policy, and network fine-tuning. Note that the following symbols are identical to those in Sec.~\ref{sec:method} of the main paper. The whole pipeline is presented in Algorithm \ref{algo:algorithm}.

\subsection{UDA warm-up} For the two tasks, GTA5 $\rightarrow$ Cityscapes and SYNTHIA $\rightarrow$ Cityscapes, we utilized a conventional UDA method \cite{tsai2018learning} to train an initial model. For both DeepLabV2 and DeepLabV3+ network backbones, we apply the SGD optimizer with an initial learning rate of 2.5e-4 and a decay rate of 0.9. Following \cite{tsai2018learning}, we warm up the network with adversarial training for about 100k steps (about 8 hours).

\subsection{Density-aware selection} As mentioned in Sec.~\ref{subsec:density} of the main paper, we utilized a set of Gaussian Mixture Models (GMMs) to model the conditional probability distributions of the two domains. The domain density $p_S(z|c)$ and $p_T(z|c)$ are the likelihood of sampling the region feature $z$ from the source domain and target domain given the predicted category $c$ respectively.

The feature $z$ is a vector of 256 dimensions. For the DeepLabV3+ network backbone, the feature is extracted before the final linear classification layer. For the DeepLabV2 network backbone, we slightly modify the network to make the output of Atrous Spatial Pyramid Pooling (ASPP) as a 256-dimensional feature vector and add the final classification layer after ASPP.

In our implementation, the number of mixtures in GMM is proportional to the number of regions of the category and is clipped in the range of 1 to 10. The process of constructing GMMs can be efficiently completed by offline and parallel execution, which takes about 0.009 seconds per region with four parallel processes. We believe the density estimator could be replaced by other methods and is worth further investigation, which is beyond the scope of this paper.

\subsection{Dynamic Scheduling Policy}

As explained in Sec.~\ref{subsec:adaptive} of the main paper, we use two hyper-parameters, $\alpha$ and $\beta$, to dynamically schedule the labeling budgets of different active selection methods. 
For the GTA $\rightarrow$ Cityscapes task, we set $\alpha = 1, \beta = 1$. For the SYNTHIA $\rightarrow$ Cityscapes task, we set $\alpha = 0.5, \beta = 1$. The selection of $(\alpha, \beta)$ is not difficult: (1) a larger $\alpha$ indicates a larger domain gap and (2) simply setting $\beta$ = 1 (half-decay scheduling) performs well on both tasks. The selection of $\alpha$ and $\beta$ (half-decay) are discussed in Sec.~\ref{sec:alpha} and Sec.~\ref{sec:beta} respectively. The computational cost for this step, including the labeling budget decision and uncertainty selection, consumes about 0.001 for each region.

\subsection{Network Fine-tuning}

After the active selection step, we acquire ground truth labels of the top-ranked regions in $D_T^U$ and move them to $D_T^L$. Then, fine-tune the model on $D_S \cup D_T^L$ using cross-entropy loss in a supervised manner. For the fine-tuning step, we utilized the SGD optimizer with an initial learning rate of 2.5e-4 and a decay rate of 0.9. The fine-tuning step takes about 8 hours.

\begin{algorithm}

\SetAlgoLined
\KwIn{data pool: ($D_S, D_T = \{D_T^L, D_T^U\}$, where $D_T^L = \emptyset$), hyper-parameters: ($\alpha, \beta$), $\rm max\_active\_iterations = N$, labeling budget for each active selection round: $B$}
\KwOut{the output model $h_\theta$}
\texttt{\\}
$\textbf{Warm-up: } h_\theta \leftarrow (D_S, D_T^U, D_T^L)$ according to \cite{tsai2018learning} \\
 \For{$n \leftarrow$ $0$ \bf{to} $N$}{
    \tcp{Construct source and target density estimators}
	$Z_S, C_S \leftarrow h_\theta(D_S)$ \\
	$Z_T, C_T \leftarrow h_\theta(D_T)$ \\
	$GMM_{\rm src} \leftarrow \textsc{ConstructGMMs}{(Z_S, C_S)}$ \\
	$GMM_{\rm trg} \leftarrow \textsc{ConstructGMMs}{(Z_T, C_T)}$ \\
    \texttt{\\}
    
    \tcp{Calculate the region importance metric $\pi$}
    $R^* \leftarrow$ a new empty list \\
    \For{\bf{each} $c$ in $\{1, 2, ..., C\}$}{
		$R_c \leftarrow $ Obtain all regions predicted as category $c$ in $D_T^U$ \\
		\For{\bf{each} $\rm region$ in $R_c$}{
		    Obtain the domain density $d_S, d_T$ by feeding the regional feature $z$ and the predicted category $c$ to $GMM_{\rm src}$ and $GMM_{\rm trg}$ \\
		    Calculate the importance score $\pi$ according to Eq. 2
		}
		Rank all the regions in $R_c$ in descending order based on the important score; then, append it to $R^*$ \\
		Calculate the categorical KL-divergence $D_{KL}(p_T (z|c)\ ||\ p_S (z|c))$ according to Eq. 5
    }
    \texttt{\\}
    
    \tcp{Dynamic Scheduling Policy}
    Determine $B_n^u, B_n^d$ given $(\alpha, \beta, B)$ according to Eq. 7 \\
    \tcp{Class-Balanced Selection}
    Determine $B^{d,c}$ for each $c$ given the categorical KL-divergence according to Eq. 6 \\
    \texttt{\\}
    
	\tcp{Label Acquisition}
	$S^u_n$ $\leftarrow$ \textsc{UncertaintySelection}($B_n^u$, $h_\theta, D_T^U$) according to \cite{ijcnn} \\
	$S^d_n$ $\leftarrow$ \textsc{DensityAwareSelection}($B^d_n$, $R^*$) { \tcp {Pick top-ranked regions based on the categorical budgets $B^{d,c}$}}
	\texttt{\\}
	
	$X_{\rm{active}} \leftarrow S^u_n \bigcup S^d_n$ \\
	$Y_{\rm{active}} \leftarrow$ Obtain ground-truth labels from the oracle given $X_{\rm{active}}$ \\
	$D_T^L \leftarrow D_T^L \bigcup (X_{\rm{active}}, Y_{\rm{active}})$ \\
	$D_T^U \leftarrow D_T^U \setminus X_{\rm{active}}$ \\
    \texttt{\\}
    
	\tcp{Supervised fine-tuning}
	$h_\theta$ $\leftarrow$ Tune the model on $D_S \bigcup D_T^L$ with cross-entropy loss
 }
 \Return $h_\theta$
 \caption{The pipeline of \ourAbbrname{}}
 \label{algo:algorithm}
\end{algorithm}

\section{Baseline Active Learning Methods}
\label{sec:al}

We describe the implementation of 8 region-based active learning baselines used in our experiments. In the following section, we let $R$ denote a region with $N$ pixels within it, and $\theta$ denotes the fixed trained deep learning network.

\subsubsection{RAND} Randomly select few regions of images in the unlabeled target domain dataset $D_T^U$ for label acquisition.


\subsubsection{MAR \cite{ijcnn}} 

Wang~\etal \cite{ijcnn} proposed using model softmax margin as the indicator to select informative instances for labeling. Specifically, we produce the score for a region ($S_R^{\text{MAR}}$) by averaging the difference between the two most likely category labels for all pixels within the region, as shown in Eq.~\ref{eq:mar}. After that, we acquire the ground truth labels of few regions with the smallest score, which means the smallest margin, in the unlabeled dataset for each category.

\begin{equation}
    S_{R}^{\text{MAR}} = \frac{1}{N} \sum_{n=1}^N P(\hat{y_n^1} | R;\theta) - P(\hat{y_n^2} | R;\theta),
    \label{eq:mar}
\end{equation}
where $\hat{y_n^1}$ is the first most probable label category and $\hat{y_n^2}$ is the second most probable label category.

\subsubsection{CONF \cite{ijcnn}} The main concept of the confidence selection strategy is to acquire labels for samples whose prediction has the least confidence \cite{ijcnn,wang2016cost}. As can be observed in Eq. \ref{eq:conf}, the score for a region ($S^{\text{CONF}}_R$) is produced by averaging the softmax confidence of all pixels within the region. After that, we select a portion of regions with the least confidence score in the unlabeled dataset for label acquisition.

\begin{equation}
    S^{\text{CONF}}_R = \frac{1}{N} \sum_{n=1}^N P(\hat{y_n^1} | R;\theta),
    \label{eq:conf}
\end{equation}
where $\hat{y_n^1}$ is the softmax confidence value of the predicted category label.

\subsubsection{ENT \cite{ijcnn}} In the field of information theory, entropy is a widely used metric to evaluate the information of a probability distribution \cite{shannon1948mathematical}. The idea of this type of selection strategy is to select regions with the largest entropy for labeling \cite{ijcnn}. As shown in Eq. \ref{eq:ent}, the score for a region ($S^{\text{ENT}}_R$) is formed by averaging the softmax entropy of all pixels in a region. After that, a portion of regions with the largest entropy in the unlabeled dataset is selected for label acquisition.

\begin{equation}
    S^{\text{ENT}}_R = - \frac{1}{N} \sum_{n=1}^N\sum_{i=1}^{c} [P(y_n^i | R;\theta) \cdot \log [P(y_n^i | R;\theta)]],
    \label{eq:ent}
\end{equation}
where $c$ is the number of categories, and $P(y_n^i | R;\theta)$ represents the softmax probability that the model predicts pixel $n$ as class $i$.

\subsubsection{BADGE \cite{ash2020babdge}} Ash~\etal proposed selecting a batch of diverse and uncertain samples for labeling with a designed gradient embedding space. Specifically, the method first calculates the gradient embedding of each sample to indicate its uncertainty and then selects diverse samples for label acquisition with k-means++. In our implementation, we produce the regional gradient embedding by averaging the value of all pixels within it. Then, we follow its original implementation to cluster the regions with the k-means++ algorithm.

\subsubsection{ReDAL \cite{wu2021redal}} Wu~\etal proposed acquiring a batch of diverse point cloud regions with high uncertainty for labeling by entropy, 3D characteristics, and greedy diversity selection. In our implementation, we carefully replaced the 3D characteristics term as detected 2D edges and followed the rest of the algorithm.

\subsubsection{AADA \cite{su2020active}} Su~\etal presented the first active domain adaptation approach for image classification. The concept of this method is to leverage the softmax entropy and the domain discriminator to select uncertain samples that are far from the source domain distribution. In our implementation, we calculated the region-level softmax entropy and domain discriminator result and followed the rest of the algorithm.

\subsubsection{CLUE \cite{prabhu2021active}} Prabhu~\etal presented another active domain adaptation method by clustering the uncertainty-weighted embeddings. The same, we treated a region as the fundamental label query unit and followed the original implementation.

\section{More Experimental Results and Analyses}
\label{sec:exp}

In this section, we first discuss the effectiveness of UDA warm-up and hyper-parameter selections. Then, we analyze the influence of inaccurately predicted categories. Finally, we report the raw tables of Fig.~\ref{fig:ALs} in the main paper in Tab.~\ref{tab:gta}, \ref{tab:synthia} and show the per-class performance of our active learning strategy in Tab.~\ref{tab:active_raw}.



\subsection{Effectiveness of UDA Warm-up}
\label{sec:warmup}

Tab.~\ref{tab:warmup} shows the mIoU scores after applying the UDA method \cite{tsai2018learning} as the warm-up step. The result shows that the performance of the UDA method is still far from that of full supervision and our method. 

We further investigate the effectiveness of UDA warm-up on the GTA5 $\rightarrow$ Cityscapes task. With 1\% target domain labeled regions, our method can reach $64.0 \sim  64.1$ mIoU with and without warm-up. This shows that UDA warm-up plays little role in our improvement. The main reason for using warm-up in our experiments is to follow prior domain adaptation works \cite{ning2021multi,shin2021labor,zhang2021prototypical}.

\begin{table}[!h]
\caption{mIoU scores of UDA warm-up \cite{tsai2018learning} on the two tasks.}
\centering
\begin{tabular}{ccc}
\toprule
\multicolumn{3}{c}{(a) GTA $\rightarrow$ Cityscapes}     \\
\midrule
          & DeepLabV2     & DeepLabV3+    \\
mIoU (\%)      &      44.61         &   45.51            \\
\bottomrule
\toprule
\multicolumn{3}{c}{(b) SYNHTIA $\rightarrow$ Cityscapes} \\
\midrule
          & DeepLabV2     & DeepLabV3+    \\
mIoU (\%)      &      39.95         &    43.04      \\
\bottomrule
\end{tabular}

\label{tab:warmup}
\end{table}

\subsection{Effectiveness of Initial Balance Coefficient}
\label{sec:alpha}

As mentioned in Sec.~\ref{subsec:adaptive} of the main paper, the balance coefficient $\alpha$ is designed to balance between density-aware and the uncertainty-based method at the first active selection round. We investigate the effectiveness of $\alpha$ with the DeepLabV3+ model backbone for the two tasks. 

As can be observed in Tab.~\ref{tab:alpha}, with the aid of partial or full acquired annotations by our designed density-aware method, models are able to obtain higher mIoU scores compared with only using conventional uncertainty-based method ($\alpha = 0$). For the GTA5 $\rightarrow$ Cityscapes task, only adopting density-aware selection strategy at the beginning achieve the best result; while for the SYNTHIA $\rightarrow$ Cityscapes task, choosing $\alpha = 0.5$ to combine density-aware and uncertainty-based methods obtain the best performance. The experimental results confirm the effectiveness of our density-aware selection in severe domain shift.

\begin{table}[h!]
\caption{We report the mIoU scores with different balance coefficients $\alpha$. We found that using only the uncertainty-based method, \ie, $\alpha = 0$,  obtained the worst results among all combinations. The results show that using some or all of the obtained annotations through density-aware selection can improve model performance.}
\centering
\begin{tabular}{cccccc}
\toprule
\multicolumn{6}{c}{(a) GTA $\rightarrow$ Cityscapes with 1\% Target Labels}     \\
\midrule
  $\alpha$   & 0  & 0.25  & 0.5 & 0.75 & 1.0  \\
mIoU (\%)    &  59.57   & 61.95  &  63.30 & 63.73 &     \textbf{64.03}       \\
\bottomrule
\toprule
\multicolumn{6}{c}{(b) SYNHTIA $\rightarrow$ Cityscapes with 1\% Target Labels} \\
\midrule
  $\alpha$   & 0  & 0.25  & 0.5 & 0.75 & 1.0  \\
mIoU (\%)    &  61.56   & 61.98  &  \textbf{62.47} &  62.07 &     61.87       \\
\bottomrule
\end{tabular}

\label{tab:alpha}
\end{table}

\subsection{Effectiveness of Different Scheduling Policies}
\label{sec:beta}

As discussed in Sec.~\ref{subsec:adaptive} of the main paper, due to the rapid domain shift reduction, we design a dynamic scheduling policy to half decay the labeling budget of the domain exploration and gradually put more emphasis on the uncertainty-based method. Here we discuss the effectiveness of five budget scheduling policies on the GTA5 $\rightarrow$ Cityscapes task with DeepLabV3+ model backbone, including pure density, pure uncertainty, even distribution, linear decay, and half decay.

We classify these five policies based on the $\lambda$ value in Eq.~\ref{eq:lambda} in the main paper. Pure density and pure uncertainty refer to $\lambda = 1$ and $\lambda = 0$ respectively. Even distribution means evenly assigning the labeling budgets to our density-aware method and uncertainty-based approach for each active selection round, \ie, $\lambda = 0.5$. Linear decay refers to the linear decrease of the labeling budgets assigned to the density-aware method. In our implementation, the proportion of density-aware selection method is initialized as $1.0$ and linearly decreases by $0.2$ for each step, \ie, $\lambda = 1.0 - 0.2(n-1)$. Half decay is our budget scheduling policy described in Sec.~\ref{subsec:adaptive} of the main paper, \ie, $\lambda = \alpha \cdot 2^{-\beta(n-1)}, (\alpha, \beta) = (1, 1)$. 

As shown in Tab.~\ref{tab:schedule}, our half-decay approach obtains the best performance with 1\%, 3\%, and 5\% budgets. Overall, the result suggests that our density-aware technique and traditional uncertainty-based method complement each other to reach better adaptability under our dynamic scheduling policy.

\begin{table}[h!]
\caption{We compare different label budget scheduling strategies on the GTA $\rightarrow$ Cityscapes task. The result shows that our designed half-decay method performs the best among all strategies.}
\centering
\begin{tabular}{c|ccc}
\toprule
\multirow{2}{*}{\begin{tabular}[c]{@{}c@{}}Scheduling Policy\end{tabular}} & \multicolumn{3}{c}{mIoU} \\
& 1\%    & 3\%    & 5\%    \\
\midrule
Pure Density ($\lambda = 1$)  &   \textbf{64.03}     &   69.38     &   70.69     \\
Pure Uncertainty ($\lambda = 0$)   &   59.97    &   68.79     &   70.70     \\
Even Distribution ($\lambda = 0.5$)  &   63.30     &   69.66     &   71.15     \\
Linear Decay  ($\lambda = 1.0 - 0.2(n-1)$)      &  \textbf{64.03}      &    69.49    &   71.14     \\
 \textbf{Our Half Decay}  ($\lambda = 2^{-(n-1)}$)       &   \textbf{64.03}     &    \textbf{69.86}  &  \textbf{71.25}   \\
\bottomrule
\end{tabular}

\label{tab:schedule}
\end{table}

\subsection{Influence of Inaccurately Predicted Categories} As mentioned in Sec.~\ref{subsec:density} in the main paper, our density-aware selection estimates the domain density with the extracted region features and the corresponding predicted categories. Since the predicted category might be inaccurate, especially in the first stage in the ADA, we conducted an experiment to verify whether our method is robust under this issue.

The result shows that indeed the initially predicted category might indeed be inaccurate, but our method can recall more of these mispredicted data for labeling. According to the statistics, about 56\% of ``target bus regions" were predicted as other classes by the initial model. Still, our method recalled 25\% of these mispredicted regions to re-label, while the uncertainty-based methods could only recall 4\% of them. Overall, we show our approach is effective even with noisy initial labels in this experiment.

\begin{table*}[h!]
    \caption{Results of mIoU performance (\%) on GTA \cite{richter2016GTA5} $\rightarrow$ Cityscapes \cite{cordts2016cityscapes} with DeepLabV3+ network backbone.}
    \centering
    \begin{tabular}{c|ccccccccc}
         \% Target Labels & RAND & MAR & CONF & ENT & BADGE & ReDAL & AADA & CLUE & \textbf{\ourAbbrname{}} \\
         \hline \hline
1 & 58.81 & 57.95 & 59.90 & 59.57 & 62.67 & 61.97 & 55.84 & 60.13 & \textbf{64.03} \\
2 & 61.11 & 64.33 & 65.72 & 66.08 & 65.62 & 65.93 & 61.71 & 62.96 & \textbf{67.65} \\
3 & 62.23 & 67.08 & 68.83 & 68.79 & 67.73 & 67.12 & 64.93 & 64.99 & \textbf{69.86} \\
4 & 63.62 & 68.47 & 69.55 & 69.83 & 68.66 & 68.55 & 65.74 & 65.88 & \textbf{70.66} \\
5 & 63.50 & 69.44 & 70.70 & 70.70 & 70.50 & 68.30 & 66.01 & 67.13 & \textbf{71.25}
    \end{tabular}

    \label{tab:gta}
\end{table*}

\begin{table*}[h!]
    \caption{Results of 16-classes mIoU performance (\%) on SYNTHIA \cite{ros2016synthia} $\rightarrow$ Cityscapes \cite{cordts2016cityscapes} with DeepLabV3+ network backbone.}

    \centering
    \begin{tabular}{c|ccccccccc}
         \% Target Labels & RAND & MAR & CONF & ENT & BADGE & ReDAL & AADA & CLUE & \textbf{\ourAbbrname{}} \\
         \hline \hline
1 & 59.05 & 60.67 & 61.94 & 61.56 & 61.64 & 61.08 & 54.16 & 60.17 & \textbf{62.47} \\
2 & 60.96 & 67.23 & 68.20 & 67.75 & 66.29 & 66.58 & 59.84 & 64.71 & \textbf{69.35} \\
3 & 63.65 & 69.68 & 69.70 & 70.23 & 69.18 & 69.30 & 63.12 & 66.50 & \textbf{71.01} \\
4 & 65.53 & 70.43 & 71.17 & 71.28 & 70.26 & 69.97 & 66.09 & 67.46 & \textbf{72.40} \\
5 & 66.09 & 71.37 & 71.50 & 71.76 & 71.08 & 71.01 & 67.03 & 68.37 & \textbf{72.74}
    \end{tabular}

    \label{tab:synthia}
\end{table*}

\begin{table*}[t]
\caption{Complete experimental results of our proposed \ourAbbrname{} on (a) GTA5 $\rightarrow$ Cityscapes and (b) SYNTHIA $\rightarrow$ Cityscapes with different percentage of acquired target labels.}
\definecolor{LightCyan}{rgb}{0.88,1,1}
\renewcommand{\arraystretch}{1.2}
\begin{center}
\centering
\resizebox{\textwidth}{!}{
\begin{tabular}{c c c c c c c c c c c c c c c c c c c c c|c}
\toprule
\multicolumn{21}{c}{(a) GTA5 $\to$ Cityscapes}\\
\midrule
\multirow{1}*{} & \% Target Labels  & Road & SW & Build & Wall & Fence & Pole & TL & TS & Veg. & Terrain & Sky & PR & Rider & Car & Truck & Bus & Train & Motor & Bike & mIoU  \\
\midrule
\multirow{5}*{\shortstack[c]{\ourAbbrname{}\\(DeepLabV2)}}
& 1\% & 93.63 & 59.90  & 86.92 & 39.59 & 40.95 & 44.04 & 51.78 & 53.87 & 88.34 & 45.30  & 86.60  & 71.09 & 46.82 & 89.81 & 57.58 & 69.71 & 58.65 & 52.54 & 68.45 & 63.45 \\
& 2\% & 95.34 & 69.08 & 88.54 & 48.24 & 49.26 & 45.21 & 54.41 & 59.61 & 89.00 & 52.73 & 90.64 & 73.09 & 50.23 & 91.20  & 69.36 & 73.00 & 59.99 & 56.03 & 69.57 & 67.61 \\
& 3\% & 96.11 & 72.52 & 88.98 & 48.33 & 50.52 & 46.42 & 55.35 & 62.08 & 89.55 & 53.86 & 90.69 & 74.11 & 52.69 & 91.47 & 67.90  & 77.01 & 65.13 & 59.20  & 70.75 & 69.09 \\
& 4\% & 96.27 & 73.91 & 89.28 & 49.03 & 52.66 & 47.12 & 56.44 & 63.54 & 89.73 & 56.52 & 91.76 & 74.49 & 53.74 & 91.66 & 68.25 & 76.29 & 62.99 & 59.08 & 71.21 & 69.68 \\
& 5\% & 96.29 & 73.57 & 89.26 & 50.01 & 52.26 & 47.94 & 56.91 & 64.65 & 89.27 & 53.94 & 92.25 & 73.91 & 52.86 & 91.84 & 69.67 & 78.87 & 62.70  & 57.65 & 71.05 & 69.73 \\
\midrule
\multirow{5}*{\shortstack[c]{\ourAbbrname{}\\(DeepLabV3+)}}
& 1\% & 93.19 & 59.06 & 87.50  & 37.95 & 43.54 & 45.43 & 53.63 & 47.59 & 88.23 & 44.72 & 89.73 & 72.04 & 48.58 & 91.11 & 63.40  & 68.98 & 58.56 & 54.88 & 68.47 & 64.03 \\
& 2\% & 95.50  & 69.38 & 88.91 & 43.63 & 50.05 & 48.77 & 56.19 & 58.97 & 89.39 & 51.66 & 90.68 & 73.94 & 51.31 & 91.65 & 66.52 & 72.15 & 58.69 & 57.48 & 70.40  & 67.65 \\
& 3\% & 96.27 & 73.91 & 89.37 & 47.65 & 52.37 & 50.12 & 57.14 & 64.29 & 89.50  & 55.64 & 91.50  & 75.03 & 53.03 & 92.28 & 69.97 & 77.16 & 63.13 & 57.35 & 71.54 & 69.86 \\
& 4\% & 96.77 & 76.58 & 89.75 & 47.28 & 53.79 & 52.33 & 57.92 & 65.41 & 89.90  & 56.69 & 92.27 & 75.31 & 53.01 & 92.09 & 68.77 & 76.43 & 67.25 & 58.82 & 72.16 & 70.66 \\
& 5\% & 96.97 & 77.83 & 89.97 & 45.98 & 55.04 & 52.74 & 58.69 & 65.80  & 90.37 & 58.94 & 92.14 & 75.69 & 54.36 & 92.26 & 69.04 & 78.01 & 68.51 & 59.05 & 72.33 & 71.25 \\
\bottomrule
\end{tabular}
}
\end{center}
\renewcommand{\arraystretch}{1.2}
\begin{center}
\centering
\resizebox{\textwidth}{!}{
\begin{tabular}{c c c c c c c c c c c c c c c c c c|c c}
\toprule
\multicolumn{19}{c}{(b) SYNTHIA $\to$ Cityscapes}\\
\midrule

\multirow{1}*{} & \% Target Labels  & Road & SW & Build & Wall & Fence & Pole & TL & TS & Veg. & Sky & PR & Rider & Car & Bus & Motor & Bike & mIoU & mIoU*\\
\midrule
\multirow{5}*{\shortstack[c]{\ourAbbrname{}\\(DeepLabV2)}}
& 1\% & 93.82 & 61.50  & 85.21 & 26.15 & 19.21 & 40.75 & 46.41 & 53.16 & 86.11 & 87.27 & 70.78 & 46.77 & 86.54 & 34.08 & 48.49 & 66.21 & 59.53 & 66.64 \\
& 2\% & 94.95 & 67.30  & 87.79 & 37.92 & 42.04 & 44.09 & 53.45 & 61.05 & 88.17 & 90.11 & 73.64 & 53.37 & 89.98 & 66.26 & 53.54 & 69.36 & 67.07 & 73.00 \\
& 3\% & 95.73 & 71.75 & 88.48 & 38.68 & 44.08 & 46.4  & 54.48 & 64.64 & 88.68 & 90.18 & 74.49 & 53.99 & 90.72 & 73.27 & 57.48 & 70.86 & 68.99 & 74.98 \\
& 4\% & 96.21 & 73.95 & 88.93 & 41.23 & 48.24 & 47.45 & 55.31 & 65.65 & 89.23 & 91.24 & 74.59 & 54.39 & 91.07 & 73.37 & 58.21 & 71.55 & 70.04 & 75.67 \\
& 5\% & 96.41 & 74.57 & 89.09 & 42.51 & 47.70  & 47.99 & 55.64 & 66.46 & 89.47 & 91.73 & 75.10  & 55.15 & 91.37 & 76.97 & 57.97 & 71.77 & 70.62 & 76.28 \\
\midrule
\multirow{5}*{\shortstack[c]{\ourAbbrname{}\\(DeepLabV3+)}}
& 1\% & 92.45 & 55.44 & 86.75 & 34.94 & 29.07 & 44.90  & 48.97 & 54.43 & 87.09 & 90.27 & 73.66 & 49.39 & 88.98 & 40.74 & 52.85 & 69.64 & 62.47 & 68.51 \\
& 2\% & 95.54 & 71.45 & 88.78 & 38.97 & 45.69 & 50.34 & 55.57 & 64.78 & 89.46 & 92.06 & 75.61 & 53.38 & 91.14 & 69.67 & 55.49 & 71.70  & 69.35 & 74.97 \\
& 3\% & 96.13 & 74.04 & 89.11 & 39.64 & 49.52 & 52.58 & 56.24 & 65.91 & 89.89 & 92.80  & 76.38 & 54.63 & 92.20  & 76.06 & 58.77 & 72.31 & 71.01 & 76.50  \\
& 4\% & 96.19 & 74.40  & 89.91 & 48.48 & 50.71 & 53.60  & 58.10  & 66.88 & 90.01 & 93.29 & 77.15 & 56.28 & 92.25 & 78.39 & 59.54 & 73.27 & 72.40  & 77.36 \\
& 5\% & 96.67 & 76.76 & 90.27 & 48.73 & 51.06 & 54.24 & 58.28 & 67.99 & 90.41 & 93.37 & 77.37 & 56.41 & 92.53 & 77.53 & 58.88 & 73.29 & 72.74 & 77.67 \\
\bottomrule
\end{tabular}
}
\end{center}

\label{tab:active_raw}
\end{table*}

\clearpage

\section{Qualitative Results}
\label{sec:qualitative}
Due to space limitations, we show the qualitative results in the supplementary material. As shown in Fig.~\ref{fig:vis_supp}, we show five inference results of different approaches for the GTA5 $\rightarrow$ Cityscapes domain adaptation task, including success and failure cases.

The first three rows present the results that our method outperforms the UDA method \cite{tsai2018learning} and the previous ADA approach \cite{su2020active}. As shown in the first row, our segmentation result is close to full supervision with clear boundaries. Compared with the prior UDA~\cite{tsai2018learning} and ADA~\cite{su2020active} practices, our method can better capture the scene structure and significantly outperforms them. The second and the third rows show that our method can better recognize hard categories, such as ``train" and ``fences" (shown on the red bounding box).

The fourth and fifth rows present two failure cases of our method. As observed from the red bounding boxes, our method performs worse than full supervision in these pictures' corners or boundary areas. However, compared with the two existing approaches, our method still achieves better results. We believe that the problem of poor performance in the pictures' corners or boundary areas may be improved through a better active selection strategy, which is worthy of further research in the future.

\begin{figure*}[!b]
    \centering
    \includegraphics[width=0.9\linewidth]{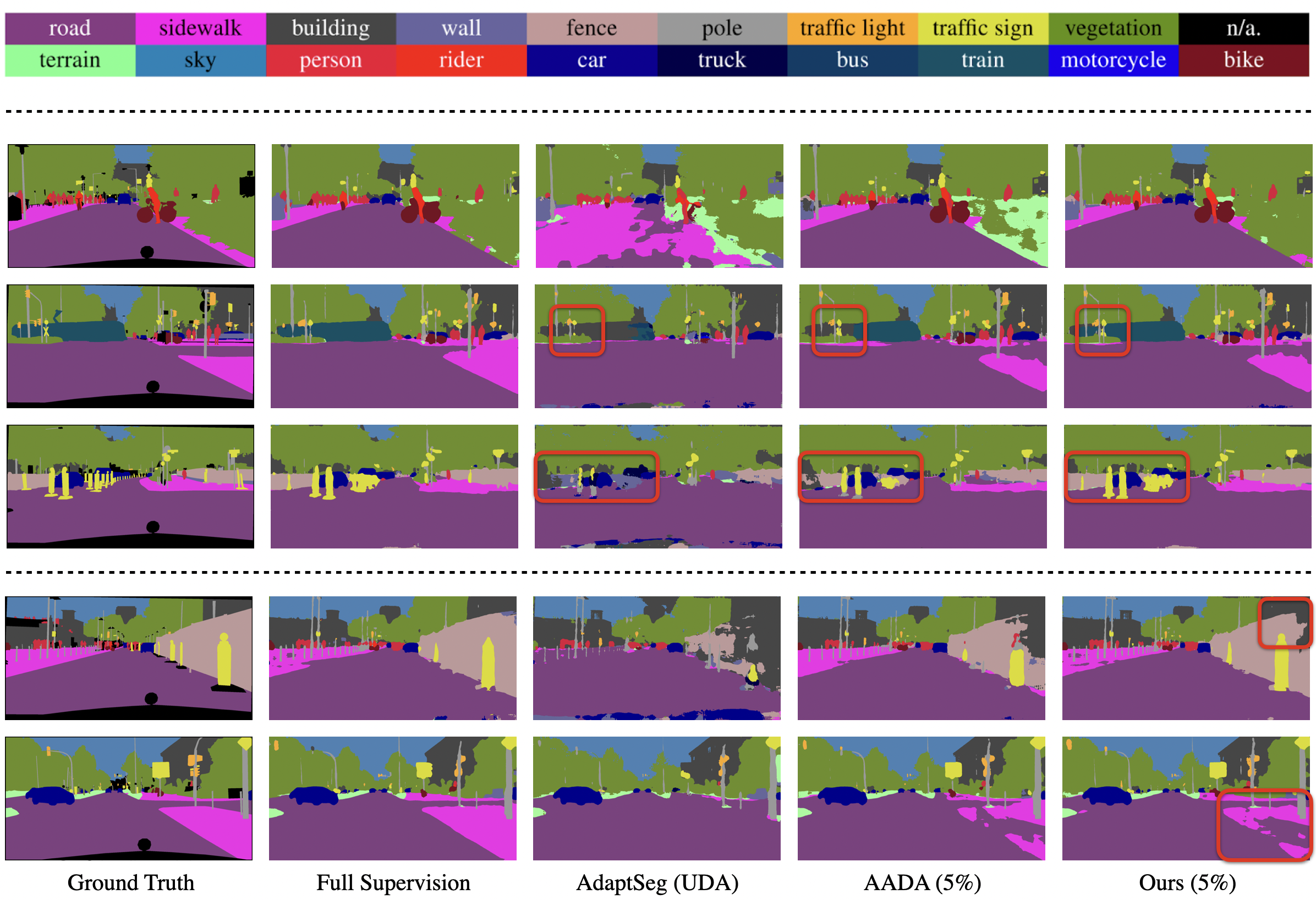}
    \caption{\textbf{Qualitative results of different approaches for the GTA5 $\rightarrow$ Cityscapes domain adaptation task.} We present three success cases (in the top three rows) and two failure cases (in the bottom two rows) of our method. For more detailed explanation, please refer to Sec.~\ref{sec:qualitative}.}
    \label{fig:vis_supp}
\end{figure*}

